\pdfoutput=1

\documentclass[11pt]{article}

\usepackage[]{acl}

\usepackage{times}
\usepackage{latexsym}

\usepackage[T1]{fontenc}

\usepackage[utf8]{inputenc}

\usepackage{microtype}

\usepackage[skip=0pt]{caption}
\usepackage{subcaption}
\usepackage{graphicx}
\usepackage{amsmath}
\usepackage{amssymb}
\usepackage{pifont}
\usepackage[dvipsnames]{xcolor}
\newcommand{\cmark}{\ding{51}}
\newcommand{\xmark}{\ding{55}}
\usepackage{booktabs}
\usepackage{algorithm}
\usepackage{algpseudocode}
\algrenewcommand\algorithmicrequire{\textbf{Input:}}
\algrenewcommand\algorithmicensure{\textbf{Output:}}
\usepackage{algpseudocode}
\usepackage{threeparttable}
\usepackage{multirow}
\usepackage{tabularx}
\usepackage{cleveref}
\usepackage{tcolorbox}
\usepackage{adjustbox}
\usepackage{soul}
\usepackage{colortbl}
\usepackage{transparent}
\usepackage{xspace}
\usepackage{paralist}
\usepackage{enumitem}

\newcommand{\customdag}{$\dag$}
\newcommand{\customddag}{$\ddag$}
\newcommand{\customdiamondsuit}{$\diamondsuit$}

\newcommand {\mm}[1] {\ifmmode{#1}\else{\mbox{\(#1\)}}\fi}

\newcommand{\interalia}{\emph{inter alia}}

\colorlet{lightgray}{lightgray!70}
\colorlet{Teal}{teal!60}
\colorlet{Magenta}{magenta!50}
\colorlet{Cyan}{cyan!40}

\crefname{figure}{Figure}{Figures}
\crefname{section}{Section}{Sections}
\crefname{equation}{Equation}{Equations}
\crefname{appendix}{Appendix}{Appendix}
\crefname{table}{Table}{Tables}

\usepackage{graphicx}

\graphicspath{{./figures/}}

\title{Multi-dimensional Evaluation of Empathetic Dialogue Responses}

\author{Zhichao Xu\textsuperscript{1}\Thanks{ Work done during ZX's internship at Google Research.} \quad
Jiepu Jiang\textsuperscript{2} \\
\\
\textsuperscript{1} Kahlert School of Computing, University of Utah \\
\textsuperscript{2} Google Research \\
{\tt  zhichao.xu@utah.edu, jiepujiang@google.com}\\
}

\begin{document}
\maketitle
\begin{abstract}

Empathy is critical for effective and satisfactory conversational communication. Prior efforts to measure conversational empathy mostly focus on expressed communicative intents---that is, the way empathy is expressed. 
Yet, these works ignore the fact that conversation is also a collaboration involving both speakers and listeners. 
In contrast, we propose a multi-dimensional empathy evaluation framework to measure both \emph{expressed intents from the speaker's perspective} and \emph{perceived empathy from the listener's perspective}. 
We apply our analytical framework to examine internal customer-service dialogues. We find the two dimensions (expressed intent types and perceived empathy) are inter-connected, while perceived empathy has high correlations with dialogue satisfaction levels.

To reduce the annotation cost, we explore different options to automatically measure conversational empathy: prompting LLMs and training language model-based classifiers. 
Our experiments show that prompting methods with even popular models like GPT-4 and Flan family models perform relatively poorly on both public and our internal datasets.
In contrast, instruction-finetuned classifiers based on Flan-T5 family models outperform prior works and competitive baselines. 
We conduct a detailed ablation study to give more insights into instruction finetuning method's strong performance.
\end{abstract}

\vspace{-5pt}
\section{Introduction}
\label{sec:intro}

Empathy as a complex socio-emotional phenomenon plays a crucial role in forming and maintaining social interactions~\citep{omdahl2014cognitive}.
Prior works in psychology and social science have highlighted the importance of empathy in human communications~\citep[][\interalia]{omdahl2014cognitive,goodwin1990conversation,goodwin2012emotion,suchman1997model}.
As speech and text-based dialogues remain prevalent in human communication, many previous studies had developed computational approaches to analyze and measure empathy expressed in dialogues~\citep[][\interalia]{rashkin-etal-2019-towards,liu-etal-2021-towards,sharma-etal-2020-computational}.

Prior efforts had proposed empathy frameworks grounded in psychology and social science theories. 
These frameworks were then adapted for specific applications or dataset domains. 
Correspondingly, researchers proposed data-driven approaches to analyze and measure empathy.
Upon reviewing existing works, we found that they mostly adopted an individualistic conceptualization of empathy~\citep{rashkin-etal-2019-towards,sharma-etal-2020-computational,welivita-pu-2020-taxonomy,welivita2021large} which positions empathy as a primarily individual mental process. These works mainly focused on modeling \emph{expressed empathy} in dialogues.
However, recent psychology research~\citep{concannon2023measuring,zaki2008takes} argued that empathy is more fundamentally relational: instead of a static personality trait, empathy is a dynamic process, aiming to serve interpersonal communicative purposes.
\citet{van2020towards} emphasized empathy as a collaborative practice, involving participation by both parties and is shaped in social interactions.
Thus, it is critical to further examine \emph{perceived empathy} in a collaborative, dynamic dialogue setting.

In this study, we propose a novel conceptual framework to evaluate empathy in multi-round dialogues.
Specifically, this framework measures both expressed and perceived empathy intents.
Our proposed multi-dimensional evaluation framework includes four perceived empathy dimensions: (1) \emph{perceived engagement}, (2) \emph{perceived understand}, (3) \emph{perceived sympathy} and (4) \emph{perceived helpfulness}.
To validate this framework's practicality, we annotated an internal dataset of 2,000 customer service dialogue sessions.
and performed a fine-grained analysis.
Our findings show a strong correlation between perceived empathy and overall dialogue satisfaction ratings, affirming the framework's efficacy in capturing empathetic traits.

The proposed empathy evaluation framework still requires subjective assessments from human experts, which are expensive and non-trivial to collect.
To scale up evaluation without excessive reliance on annotated data, we investigate two model-based evaluation approaches: prompting LLMs and training language model-based classifiers.
We show that prompting methods with popular LLMs such as GPT-4 and Flan family models perform poorly on our datasets, indicating the challenging nature of measuring conversational empathy. 
In contrast, traditional supervised fine-tuning yields better accuracy with labeled data.
Notably, instruction-finetuned classifiers based on Flan-T5 family models~\citep{chung2022scaling} achieve the best performance.
We also conduct a comprehensive ablation study of this method to provide insights into its strong performance.
To summarize, our contributions are three-fold:
\begin{itemize}
    \vspace{-5pt}
    \item We propose a novel multi-dimensional evaluation framework to measure conversational empathy from (1) expressed communicative intents and (2) perceived empathy aspects.
    \item We apply this proposed framework to analyze an internal customer service dataset. We find the two empathy dimensions are inter-connected, while perceived empathy dimension directly affects conversation satisfaction.
    \item We comprehensively study the effectiveness of different model-based evaluation methods. On both public and internal datasets, instruction-finetuned classifiers based on Flan-T5 family models outperform prompting methods and other competitive baselines.
\end{itemize}

\section{Related Work}
\vspace{-3pt}
\label{sec:related}

\noindent
\textbf{Definition of empathy in literature.}\,
Empathy has been an influential concept in ethics and moral theory~\citep{hoffman1996empathy,slote2007ethics,smith1959theory}, social/developmental psychology~\citep{hoffman1996empathy,batson2015empathy} and other domains.
Contributions from various domains result in different definitions, functions and proposed empathy components, posing a challenge to studying empathy~\citep{coplan2011understanding}.
Prior works~\citep[][\interalia]{elliott2011empathy,elliott2018therapist,althoff-etal-2016-large,sharma-etal-2020-computational} mainly consider empathy as an individual's mental process. 
Recent psychological studies highlight empathy as a collaborative practice involving participation by both parties~\citep[][\interalia]{van2020towards,yalccin2020modeling,concannon2023measuring}. 
Within this work, we adopt this interactional definition of empathy, as our aim is to measure and evaluate empathy in social scenarios such as real-time human-human conversation.

\noindent
\textbf{Measuring empathy in dialogues.}\, 
The CL community has made much effort in algorithmically generating and measuring empathetic dialog communications. For example, \citet{rashkin-etal-2019-towards} collected an empathetic dialog dataset via crowdsourcing, with each dialogue created with a preassigned emotion label and rated by participants for empathy, relevance, and fluency. 
While they explored various empathetic dialog generation models, their empathy measurement relied on BLEU~\citep{papineni2002bleu}, raising questions about the adequacy of reference-based metrics for empathy assessment. 
\citet{sharma-etal-2020-computational} focused on measuring expressed empathy intents in online mental support conversations, considering aspects like emotional reactions, interpretations, and explorations, but did not address perceived empathy.
Similarly, other works~\citep{welivita2021large,liu-etal-2021-towards} measured expressed empathy intents without considering perceived empathy, albeit with some differences in their intent categories.

Previous work on measuring empathetic dialogue is limited in two aspects. First, many studies~\citep{sharma-etal-2020-computational, welivita2021large, liu-etal-2021-towards} focused solely on expressed empathy. Although \citet{rashkin-etal-2019-towards} collected empathy ratings from the listener's perspective, their primary focus was not empathy measurement. Second, nearly all previous work~\citep[][\interalia]{sharma-etal-2020-computational, liu-etal-2021-towards} specifically focus on mental health support setting. 
Our work is different from prior works in two ways. Firstly, our measurement framework covers different forms of empathy expression (expressed empathy) and assesses how empathy is perceived by the listeners (perceived empathy). 
Secondly, we study model-based evaluation metrics on public datasets for online mental health support and an internal online customer support dataset.
This internal task-oriented dialogue dataset can provide valuable insights into empathy measurement across different contexts.

\vspace{-5pt}
\section{Empathy Measurement Framework}
\vspace{-5pt}
\label{sec:task_definition}

\textbf{Empathy as a collaborative practice.}\,
Empathy is critical in human communication for social interactions~\citep{omdahl2014cognitive}, varying in definition across psychology and social science literature~\citep[][\interalia]{hall2019empathy,yalccin2020modeling}.
\citet{elliott2018therapist} distinguishes \emph{affective} and \emph{cognitive} aspects of empathy. 
The \emph{affective} aspect relates to the emotional stimulation in reaction to the experiences and feelings expressed by others, while the \emph{cognitive} aspect is a more deliberate process of understanding and interpreting the experiences and feelings of others and communicating such understandings. 
However, this interpretation overlooks empathy's role in interactive dialogues where multiple parties engage. 
Recent psychology works empathize empathy as a collaborative, interactive process~\citep[][\interalia]{van2020towards,concannon2023measuring}, urging studies across diverse social scenarios. 
Motivated by these insights, our study adopts this interactional definition of empathy---\emph{empathy refers to an individual's ability to perceive and understand the other's emotional states and respond correspondingly.}

\noindent
\textbf{The proposed empathy evaluation framework.}\,
In our study, we define empathy as a collaborative practice involving both parties. We measure empathy in dialog interactions using a two-dimensional framework covering both \emph{expressed empathy (communicative intents)} and \emph{perceived empathy}:
\begin{itemize}[leftmargin=*]
\vspace{-5pt}
    \item \textbf{Expressed Empathy (Communicative Intent)}: we measure empathy from the speaker's perspective by predicting the specific intents of the utterance as ways to convey empathy. This dimension aligns with previous work~\citep{rashkin-etal-2019-towards,sharma-etal-2020-computational,welivita2021large}, where intents are often detailed and domain-specific.
    \item \textbf{Perceived Empathy}: We adapt the definition in recent psychology literature~\citep[][\interalia]{concannon2023measuring,yalccin2019evaluating,yalccin2020modeling} for task-oriented dialogues. We propose to assess empathy from the listener's perspective, predicting whether an utterance is perceived as empathetic. This assessment breaks perceived empathy into four fine-grained aspects:
    \begin{itemize}[leftmargin=*]
    \vspace{-3pt}
        \item \textbf{Engagement} measures the degree to which the listener perceives the speaker as being involved in the conversation. Low engagement in the conversation can lead to the listener feeling indifferent and lacking empathy.
        \item \textbf{Understanding} measures how well the listener feels the speaker understands their situation, feelings or problems. A lack of understanding can diminish perceived empathy.
        \item \textbf{Sympathy} captures the listener's perception of the speaker's ability to empathize and react appropriately. A lack of sympathy can weaken perceived empathy.
        \item \textbf{Helpfulness} determines whether the listener finds the speaker's communication to be helpful in addressing the conversation's core issues. While not directly related to empathy, in task-oriented dialogues, where addressing problems is the objective, helpfulness can affect perceived empathy.
    \end{itemize} 
\end{itemize}
\begin{table}[!t]
\centering
\caption{Comparison between previous works and our empathy measurement framework.}
\small
\resizebox{1.0\columnwidth}{!}{
\begin{tabular}{lll}
\toprule
& \begin{tabular}[c]{@{}l@{}l} Expressed \\ Empathy \\ \end{tabular} 
& \begin{tabular}[c]{@{}l@{}l} Perceived \\ Empathy \\ \end{tabular} 
\\ \midrule
\citet{rashkin-etal-2019-towards} & \xmark & \cmark (Uni-dimensional) \\
\citet{sharma-etal-2020-computational} & \cmark (3 intents) & \xmark \\
\citet{welivita2021large} & \cmark (8 intents) & \xmark \\
\citet{liu-etal-2021-towards} & \cmark (7 intents) & \xmark \\
Our framework & \cmark (16 intents) & \cmark (4 dimensions) \\
\bottomrule
\end{tabular}
}
\label{tab:difference_scheme}
\vspace{-5pt}
\end{table}

\vspace{-5pt}
The four perceived empathy aspects are inter-connected and build upon each other. \textbf{Engagement} serves as the foundation, leading to \textbf{Understanding}, which in turn forms the basis for \textbf{Sympathy} and \textbf{Helpfulness}. \textbf{Sympathy} addresses emotional needs, while \textbf{Helpfulness} addresses practical needs. Our framework, unlike previous ones, covers both expressed and perceived empathy simultaneously. It offers a multi-faceted approach to measuring perceived empathy, as summarized in Table~\ref{tab:difference_scheme}.

\begin{table*}[!t]
\centering
\caption{
Relationship between expressed intents and perceived empathy dimensions and user satisfaction (each column reports the mean value of the perceived empathy dimension's ratings when an intent appears or not); $\diamondsuit$, $\dag$, and $\ddag$ indicate a statistical significance of ratings without intent, compared to ratings with the intent, at $p<0.05$, $0.01$, and $0.001$, respectively. We report the standard deviation in Table~\ref{tab:relationship_intent_empathy_w_stderr} in Appendix.
}
\vspace{5pt}
\setlength\extrarowheight{2pt}
\begin{minipage}[t]{\linewidth}
\begin{adjustbox}{max width=\linewidth}
\begin{tabular}{l|ll|ll|ll|ll|ll}
\hline
\begin{tabular}[l]{@{}l@{}l} \large \textbf{Expressed} \\ \large \textbf{Intent} \\ \end{tabular}
& \multicolumn{2}{c|}{\begin{tabular}[l]{@{}l@{}l} \large \textbf{Perceived} \\ \large \textbf{Engagement} \\ \end{tabular} }
& \multicolumn{2}{c|}{\begin{tabular}[l]{@{}l@{}l} \large \textbf{Perceived} \\ \large \textbf{Understand} \\ \end{tabular} }
& \multicolumn{2}{c|}{\begin{tabular}[l]{@{}l@{}l} \large \textbf{Perceived} \\ \large \textbf{Sympathy} \\ \end{tabular} }
& \multicolumn{2}{c|}{\begin{tabular}[l]{@{}l@{}l} \large \textbf{Perceived} \\ \large \textbf{Helpfulness} \\ \end{tabular} }
& \multicolumn{2}{c}{\begin{tabular}[l]{@{}l@{}l} \large \textbf{Conversation} \\ \large \textbf{Satisfaction} \\ \end{tabular} }
\\ \cline{2-11}
& \large w/ INT & \large w/o INT & \large w/ INT & \large w/o INT & \large w/ INT & \large w/o INT & \large w/ INT & \large w/o INT & \large w/ INT & \large w/o INT \\ \hline
\large ask contact &   \large 3.97 &  \large 3.99  &  \large 3.90 & \large 3.96\customdag &  \large 3.05 & \large 3.08\customdiamondsuit &  \large 3.83    & \large 3.87\customdiamondsuit &  \large 3.72    & \large 3.74  \\
\large ask details                      &  \large 4.01     &  \large 3.98   & \large  3.95    &  \large 3.95  &  \large 3.07   & \large 3.08 &  \large 3.89    & \large 3.86  &  \large 3.75    & \large 3.74  \\
\large ask confirm                      & \large   4.00      &  \large  3.99  &  \large  3.96    & \large  3.95   &  \large  3.06    & \large  3.07  &  \large  3.92    & \large  3.86\customdiamondsuit   &  \large  3.85  & \large  3.73\customdag   \\
\large aware problem                    & \large   4.03     &  \large  3.97\customddag &  \large  4.03   &  \large  3.91\customddag &  \large  3.12   & \large  3.06\customddag &  \large  3.92   & \large  3.84\customddag  & \large   3.85    & \large  3.69\customddag \\
\large describe problem                 & \large   4.04     & \large   3.98\customddag &  \large  4.02   & \large  3.94\customddag &  \large  3.15    & \large  3.06\customddag  &  \large  3.93   & \large  3.85\customddag   &  \large  3.88    & \large  3.72\customddag \\
\large express sympathy                 & \large   4.07      &  \large  3.98\customddag &   \large 4.01    & \large  3.94\customdag &   \large 3.48  &  \large 3.04\customddag &   \large 3.87    &  \large 3.86   &   \large 3.82  &  \large 3.74  \\
\large express reassurance &  \large  4.06     & \large    3.97\customddag & \large   4.01   & \large  3.94\customddag & \large  3.17    & \large  3.05\customddag  & \large   3.93    &  \large 3.85\customddag  &   \large 3.84    & \large  3.72\customddag\\
\large express apology                  &  \large  3.97      &  \large  3.99 &  \large  3.96    & \large  3.95 & \large  3.12    &  \large 3.07\customdiamondsuit &   \large 3.84    &  \large 3.87  &   \large 3.69    &  \large 3.75 \\
\large answer question                  &  \large  4.08      &  \large  3.98\customddag &   \large 4.05    & \large  3.94\customddag &  \large  3.08    & \large  3.07  &  \large  4.02    & \large  3.85\customddag  &  \large  3.92  & \large  3.73\customddag \\
\large clarify &  \large  4.06    &  \large  3.99   &  \large  4.11  & \large  3.95\customddag  &  \large  3.14  & \large  3.07 &  \large  4.06  & \large  3.86\customddag & \large   3.99  & \large  3.74\customdag \\
\large explain                          & \large   4.09      &  \large  3.99\customdag &   \large 4.08    & \large  3.95\customddag &  \large  3.03    & \large  3.08 & \large   4.07    & \large  3.86\customddag &  \large  4.00  & \large  3.73\customddag \\
\large excuse                           & \large   3.96    & \large   3.99  &  \large  3.97  &  \large 3.95 &  \large  3.08  & \large  3.07 &  \large  3.78  &  \large 3.87  &  \large  3.56  & \large  3.75\customddag  \\
\large inform action                    & \large   4.06      & \large   3.98\customddag  &   \large 4.03    &  \large 3.94\customddag &  \large  3.08    &  \large 3.07   &   \large 3.98    &  \large 3.85\customddag &   \large 3.90    &  \large 3.72\customddag \\
\large instruct action                  & \large   4.02      &  \large  3.98\customdiamondsuit   & \large   3.97    & \large  3.95 &  \large  3.08    & \large  3.07  &\large 3.91    &  \large 3.86\customdiamondsuit   & \large   3.75    &  \large 3.74  \\
\large tentative solution               & \large   4.10    &  \large  3.98\customdiamondsuit  &  \large  4.06  & \large  3.95\customdag &  \large  3.16  &  \large 3.07\customdiamondsuit &  \large  4.00  &  \large 3.86\customdag    &  \large  3.87  & \large  3.74\customdiamondsuit\\
\large contact other                    & \large   3.95      &  \large  3.99 &  \large  3.88    & \large  3.95\customdiamondsuit &  \large  3.05    & \large  3.08  &  \large  3.74  & \large  3.87\customdag &  \large  3.66  & \large  3.75  \\

\hline
\end{tabular}
\end{adjustbox}
\end{minipage}
\label{tab:relationship_intent_empathy}
\vspace{-10pt}
\end{table*}

\vspace{3pt}
\noindent
\textbf{Applying the proposed framework to analyze real-world dataset.}\,
Our framework can reveal nuanced connections between the two dimensions in measuring dialogue empathy. Table~\ref{tab:relationship_intent_empathy} reports results from an internal commercial dataset on online customer support dialogues. We annotated this dataset using our two-dimensional framework (annotation details in~\cref{subsec:datasets} and~\cref{appendix:detailed_annotation}). 
First, a UX researcher identified 16 different intents using a grounded theory approach based on manual analysis of 100 conversations. Human raters then annotated these 16 expressed intents and 4 perceived empathy aspects for utterances sampled from 2,000 conversations (we sampled one utterance for each conversation and judged that utterance only). 
For each utterance, raters assessed intent presence (yes/no) and each of the 4 perceived empathy aspects using a Likert-scale from 1 to 5, where higher ratings indicate higher degrees. 
They also annotated user satisfaction for the whole conversation if they were the customer. 
We acknowledge that raters, as a third-party to these conversations, may not perfectly assess perceived empathy dimensions, but direct feedback from the customers also has limitations such as response biases.

Results in Table~\ref{tab:relationship_intent_empathy} show that the two dimensions are connected but sufficiently distinct. 
We report the ratings of the four perceived empathy aspects (each column) when each intent (each row) appears or not (``w/ INT'' and ``w/o INT''), and test whether ratings are significantly different by intent occurrence. 
First, we observed that some expressed empathy intents lead to significantly higher levels of perceived empathy in certain aspects, suggesting the value of conveying such intents to improve conversational empathy. 
Second, each perceived empathy aspect is connected with a distinct set of expressed intents and most expressed communicative intents are only related to a subset of the four perceived aspects. 
Lastly, perceived empathy aspects differ from overall user satisfaction, yet have Spearman correlation coefficients of 0.410, 0.396, 0.099 and 0.580, all significant at 0.0001 level except perceived sympathy---indicating high correlation.
In summary, these findings suggest the nuanced and multi-faceted nature of empathy measurement: two dimensions (expressed intents and perceived empathy aspects) are related but sufficiently different, and perceived empathy aspects are directly related to conversation satisfaction.

\section{Empathy Measurement Models}

This section describes model-based metrics to measure conversational empathy. Similar to previous studies~\citep[][\interalia]{sharma-etal-2020-computational,lee2022does,kim-etal-2021-perspective}, we leverage language models to build classifiers to distinguish between empathetic and non-empathetic utterances in dialogues. 

Measuring conversational empathy is an intrinsically challenging task because: 
(1) empathy is often communicated implicitly~\citep{suchman1997model,li-etal-2023-understanding,concannon2023measuring}, posing challenges for state-of-the-art language models~\cite{ruis2022large}; 
(2) empathy expression styles and intents are often domain specific; and 
(3) the nuance between expressed and perceived empathy is often subtle and depends on the two parties' social roles~\citep[][\interalia]{suchman1997model,deppermann2011study,kupetz2014empathy}. 
For example, asking clarification questions may be perceived as empathetic in a dialogue between a therapist and a patient, but can be neutral in task-oriented dialogues such as customer service dialogues.

\subsection{Problem Definition and Notations}
\label{subsec:problem_definition}
We address conversational empathy measurement as a classification problem.
Denote a dialogue as $\mathcal{D}$ consisting of a few utterances $u_i$, $\mathcal{D}=\lbrace u_1, u_2, \ldots u_{|\mathcal{D}|}\rbrace$,
we aim to learn a function $f_{\theta}: U \rightarrow Y$ parameterized by $\theta$ that maps the $i$-th utterance $u_i \in \mathcal{D}$ to a corresponding label $y$ in the label space $Y$.
Such labels can be the presence of expressed communicative intents or perceived empathy aspects.
Given the limitation of the maximum input length of language models, we use a sliding approach similar to existing works~\citep{cao-etal-2019-observing,welivita2021large}. 
In practice, function $f$ takes as input the targeted utterance $u_i$ together with its $j$ preceding utterances and $j'$ proceeding utterances as a concatenated natural language sequence, i.e. $\texttt{Concat}( u_{i-j}, \ldots, u_{i-1}, u_{i}, u_{i+1}, \ldots, u_{i+j'} )$. 
Such decision choice echos the interactional empathy interpretation---the meaning and the effect of an utterance (whether it is perceived as empathetic or not) depends on its context with both the speaker and the listener, alongside their social roles.

\begin{table*}[t]
\centering
\caption{An example illustration of the natural language instruction schema. This example is from Empathy Mental Health (EMH) dataset~\citep{sharma-etal-2020-computational} with the communicative intent of emotional reactions.}
\vspace{0pt}
\resizebox{\textwidth}{!}{
\begin{tabular}{ll}
\toprule
\begin{tabular}[c]{@{}l@{}l} Emotional \\ Reactions \\ \end{tabular} 
& \begin{tabular}[c]{@{}l@{}l@{}l@{}l@{}l@{}l@{}l@{}l@{}l@{}l}
You are a crowdsourcing annotator. Now read the following definition and \colorbox{Teal}{a corresponding dialogue between an emotional}\\
\colorbox{Teal}{support seeker and an emotional supporter provider}, then answer the corresponding question.\\ 
\colorbox{lightgray}{Definition: emotional reaction is a common strategy in online mental support. It refers to the supporter expressing emotions}\\
\colorbox{lightgray}{such as warmth, compassion, and concern about what was experienced by the help seeker after listening to the help seeker's }\\
\colorbox{lightgray}{utterance in a dialogue. A weak communication addresses those emotions without explicit reference, e.g. 'Everything will}\\ 
\colorbox{lightgray}{be fine', while strong communication specifies the emotion, e.g. 'I feel really sorry for you.'}\\
\{Dialogue\} \\
Question: \colorbox{Magenta}{by saying \{utterance\}, what is the extent of emotional reactions expressed by the supporter?} \\
\colorbox{Cyan}{Respond with strong, weak or no communication}. \\
\end{tabular} \\ 
\bottomrule
\end{tabular}
}
\label{tab:ni_schema}
\vspace{-10pt}
\end{table*}

\subsection{Language Model-based Classifiers}
\label{subsec:classifier}
\textbf{Supervised finetuning.}\,
We include four encoder models: BERT-Large~\cite{devlin-etal-2019-bert} and RoBERTa-Large~\cite{liu2019roberta}, in addition to the encoder part of Flan-T5~\cite{chung2022scaling}: Flan-T5-Enc-\{Large, XL\}.
We randomly initialize a classification head (linear layer) and finetune on labeled training set. 

For instruction-finetuning methods, we map the labels to natural language verbalizers, and instruction finetune the model to predict corresponding verbalizers. We include two Flan-T5 models: Flan-T5-\{Large, XL\}.
We design the natural language instructions to include the following parts:
(1) \colorbox{Magenta}{\textbf{intent}} describes what tasks we want the language model to perform. In our case, the task is to predict one specific utterance's expressed communicative intent or perceived empathy.
(2) \colorbox{lightgray}{\textbf{definition}} describes the definition of the communicative intent we want the language model to predict and can be skipped if the communicative intent is simple and self-explanatory.
(3) \colorbox{Teal}{\textbf{domain}} describes the dialogue domain, such as mental therapy or customer service dialogue.
(4) \colorbox{Cyan}{\textbf{options}} refers to what options (verbalizers) should the language model predict from, such as an expressed empathy intent appears or not.
\cref{tab:ni_schema} shows an example of such natural language instruction schema. 
Complete instruction tuning templates are given in~\cref{appendix:prompt_template}.

\noindent
\textbf{Prompting method.}\,
We use a similar prompting format as instruction finetuning methods, and experiment with two open source instruction finetuned encoder-decoder language models---Flan-T5-XXL~\cite{chung2022scaling} and Flan-UL2~\cite{tay2022unifying}. 
We also use a proprietary model \texttt{GPT-4-0613}~\cite{openai2023gpt}.
For GPT-4, we use zero-shot and few-shot prompting while for Flan-T5-XXL and Flan-UL2, we only use zero-shot due to their limited 512 context length.

\section{Experimental Setup}
\label{sec:experiment_setup}
\subsection{Datasets}
\label{subsec:datasets}

We use two public empathy dialogue understanding datasets and our internal dataset. Detailed datasets statistics are given in~\cref{appendix:datasets} and~\cref{tab:dataset_statistics}.

\noindent
\textbf{Empathy Mental Health (EMH)}~\citep{sharma-etal-2020-computational} is collected from mental health-focused subreddits and consists of 3.1k single-round, asynchronous dialogues on online mental health support.
This dataset is a multi-label classification dataset where each dialogue is annotated on 3 empressed empathy dimensions at 3 levels.

\noindent
\textbf{Emotion Support Conversation (ESConv)}~\citep{liu-etal-2021-towards} is collected with crowdworkers chatting in help-seeker and supporter roles, to simulate the scenario of synchronous peer-to-peer mental counseling and support.
The dataset consists of 1.3k multi-round dialogues and is annotated with 3 mental support stages and 7 fine-grained mental support strategies. The dataset was originally designed for multi-class classification for the 7 communicative strategies (intents), and we adapt it to multi-label classification setting.

\noindent
\textbf{Empathy Evaluation (Empeval)} is our internal dataset (due to privacy concerns, we cannot release the annotated dataset, but we try to present a holistic view to assist reproducibility). 
This dataset includes 1,833 anonymized conversations sampled from a company's customer support logs, where the conversations are between customers and human representatives. All personal identifiers have been removed. On average, each conversation has 41.4 utterances, and each utterance has 7.4 tokens measured by T5 models' WordPiece tokenizer. When we built the dataset, we randomly sampled 2,000 conversations from the log and used an internal human annotation platform to judge the conversations. The annotation questions are provided in~\cref{tab:annotation_questions_empeval_p1,tab:annotation_questions_empeval_p2}. 
Each conversation were annotated by two different raters recruited from an internal annotation platform. 
167 non-English conversations were discarded. On average, the two raters have achieved 90.23\% agreement rate and 0.615 Cohen's kappa on the judgments---indicating a moderate rating consistency. 
For expressed communicative intents, we mark the intent as \texttt{True} when both annotators mark it as True. 
For the four perceived empathy dimensions, we convert the numerical annotations to binary labels by marking as \texttt{True} when both annotators give a score higher than 4. 

For all three datasets, we use five different train-test splits of 80-20 on the dialogue level, then we split the dialogues into utterances and conduct utterance-level training and predictions.
We use the standard practice for multi-label classification problems ---for each label, we train a classifier. 
Therefore for each fold we train 3, 7 and 20 classifiers for EMH, ESConv and Empeval, respectively, the reported results are averaged over 5 splits.

\subsection{Compared Methods}
\label{subsec:baselines}

\textbf{Prior methods.}\,
A few prior studies aim to design model-based metrics to measure conversational empathy, we include following methods:

\noindent
\textbf{\citet{sharma-etal-2020-computational}} use a bi-encoder model initialized with RoBERTa-Base~\citep{liu2019roberta}, to jointly predict listener's communicative intent and extract rationales. 
The method is designed for single-round conversations thus we also report the single-round performances on our two multi-round dialogue datasets ESConv and Empeval.

\noindent
\textbf{\citet{welivita2021large}} use a customized structure based on RoBERTa-Base~\cite{liu2019roberta}. The model additionally uses an utterance level attention to aggregate hidden states of utterances within a dialogue window to be used by the classifier.

\noindent
\textbf{\citet{li-etal-2023-understanding}} uses a sentence-level encoder method based on RoBERTa-Large with continued in-domain pretraining as~\citep{gururangan-etal-2020-dont}.

\noindent
\textbf{Language model-based classifiers.}\,
As previously discussed, we include six supervised finetuning methods: BERT-Large, RoBERTa-Large, Flan-T5-Enc\{Large, XL\}, Flan-T5-\{Large,XL\}.

\noindent
\textbf{Prompting methods.}\, We include Flan-T5-XXL Zero-shot, Flan-UL2 Zero-shot, GPT-4 \{Zero-shot, Few-shot\}.
For GPT-4 Few-shot, we use a prompt template where each class has one in-context example, i.e. 3-shot for EMH dataset and 2-shot for ESConv.
Due to privacy reason, we did not conduct GPT-4 experiments on internal Empeval dataset.

\subsection{Evaluation Metrics}
We report Macro Precision, Macro Recall, Macro F1 and accuracy for each prediction task.
For each dataset, we report the mean values of each metric across different communicative intents/strategies, full results are given in~\cref{subsec:full_results}.

\subsection{Implementation Details}
\textbf{Instruction and prompt templates.}\,
One of the authors manually writes the natural language instructions used for instruction finetuning and the prompt templates for prompting methods. 
We experiment with three different sets of prompting templates for prompting method, and select the performant one on a held-out development set of 100 instances. 
All the templates for instruction finetuning and prompting are given in~\cref{appendix:prompt_template}.

\noindent
\textbf{Loss functions.}\,
\label{subsec:loss_fct}
We notice longtail distributions of labels on all three datasets, due to the nature of human-human dialogues and categorization of communicative empathy.
In addition to the standard cross-entropy loss function, we additionally include two other loss functions: Focal loss~\citep{lin2017focal} and LDAM loss~\cite{cao-etal-2019-observing} to tackle class imbalance problem.
The main performance reported is still based on standard cross-entropy loss and we use these two additional loss functions as an ablation study (detailed in Section~\ref{subsec:ablation}).

\begin{table*}[t]
\centering
\caption{Main results on three datasets. We mark the best number within each section as \textbf{bold} and denote the best number in each column with $\dag$. Full results are referred to~\cref{subsec:full_results}.}
\vspace{0pt}
\setlength\extrarowheight{1pt}

\resizebox{1.0\linewidth}{!}{
\begin{tabular}{l|llll|llll|llll}
\toprule
\textbf{Datasets} & \multicolumn{4}{c|}{\textbf{EMH}}  & \multicolumn{4}{c|}{\textbf{ESConv}} & \multicolumn{4}{c}{\textbf{Empeval}} \\ 
\hline
\textbf{Metrics} 
& \begin{tabular}[c]{@{}l@{}l} \textbf{Macro} \\ \textbf{Pre.} \\ \end{tabular} 
& \begin{tabular}[c]{@{}l@{}l} \textbf{Macro} \\ \textbf{Rec.} \\ \end{tabular} 
& \begin{tabular}[c]{@{}l@{}l} \textbf{Macro} \\ \textbf{F1} \\ \end{tabular} 
& \begin{tabular}[c]{@{}l@{}l} \textbf{ACC}. \\ \, \\ \end{tabular} 
& \begin{tabular}[c]{@{}l@{}l} \textbf{Macro} \\ \textbf{Pre.} \\ \end{tabular} 
& \begin{tabular}[c]{@{}l@{}l} \textbf{Macro} \\ \textbf{Rec.} \\ \end{tabular} 
& \begin{tabular}[c]{@{}l@{}l} \textbf{Macro} \\ \textbf{F1} \\ \end{tabular} 
& \begin{tabular}[c]{@{}l@{}l} \textbf{ACC}. \\ \, \\ \end{tabular} 
& \begin{tabular}[c]{@{}l@{}l} \textbf{Macro} \\ \textbf{Pre.} \\ \end{tabular} 
& \begin{tabular}[c]{@{}l@{}l} \textbf{Macro} \\ \textbf{Rec.} \\ \end{tabular} 
& \begin{tabular}[c]{@{}l@{}l} \textbf{Macro} \\ \textbf{F1} \\ \end{tabular} 
& \begin{tabular}[c]{@{}l@{}l} \textbf{ACC}. \\ \, \\ \end{tabular} \\ 
\rowcolor{lightgray}
\multicolumn{13}{l}{\textsc{Prior methods}} \\
\textbf{\citet{sharma-etal-2020-computational}} & 72.5 & 70.7 & 71.4 & 85.8 & \textbf{83.2} & 68.1 & 70.7 & \textbf{90.9} & 70.1 & 68.4 & 68.6 & 88.4 \\
\textbf{\citet{welivita2021large}} & 68.5 & 68.7 & 67.8 & 85.2 & 80.2 & \textbf{72.9} & \textbf{74.6} & 90.8 & 70.7 & 69.7 & 69.9 & 89.1 \\
\textbf{\citet{li-etal-2023-understanding}}  & \textbf{75.3} & \textbf{70.9} & \textbf{72.7} & \textbf{86.3} & 78.8 & 72.1 & 74.4 & 90.8 & \textbf{72.7} & \textbf{70.7} & \textbf{71.0} & \textbf{89.5} \\
\rowcolor{lightgray}
\multicolumn{13}{l}{\textsc{Encoder Models}} \\
\textbf{BERT-Large} & 76.9 & 73.8 & 75.1 & 87.5 & 75.5 & 71.7 & 73.1 & 90.8 & 76.7 & \textbf{74.6} & \textbf{75.9} & 91.1 \\
\textbf{RoBERTa-Large} & \textbf{77.5} & 75.8 & 76.0 & 87.5 & 71.8 & 71.4 & 71.4 & 90.6 & \textbf{78.5} & 73.8 & 75.5 & \textbf{91.5} \\
\textbf{Flan-T5-Enc-Large} & 73.7 & 74.8 & 74.1 & \textbf{88.5} & \textbf{77.0} & 71.9 & 72.8 & \textbf{90.9} & 74.0 & 71.4 & 72.6 & \textbf{91.5} \\
\textbf{Flan-T5-Enc-XL} & 77.4 & \textbf{76.1} & \textbf{76.6} & 88.4 & 76.7 & \textbf{73.1} & \textbf{73.5} & \textbf{90.9} & 78.3 & 74.5 & 75.4 & 90.7 \\
\rowcolor{lightgray}
\multicolumn{13}{l}{\textsc{Instruction Finetuned Encoder-decoder Models}} \\
\textbf{Flan-T5-Large} & \textbf{78.7}$\dag$ & 74.1 & 74.8 & 88.2 & 76.8 & \textbf{74.8}$\dag$ & \textbf{75.3}$\dag$ & 91.2 & 80.5 & 75.8 & 77.2 & 92.1 \\
\textbf{Flan-T5-XL} & 77.8 & \textbf{76.6}$\dag$ & \textbf{77.0}$\dag$ & \textbf{88.9}$\dag$ & \textbf{78.6}$\dag$ & 73.5 & 75.1 & \textbf{91.6}$\dag$ & \textbf{81.7}$\dag$ & \textbf{76.8}$\dag$ & \textbf{78.3}$\dag$ & \textbf{92.5}$\dag$ \\
\rowcolor{lightgray}
\multicolumn{13}{l}{\textsc{Prompting Methods}} \\
\textbf{Flan-T5-XXL Zero-shot} & 45.5 & 49.3 & 42.9 & 62.6  & 62.8 & 71.8 & 61.5 & 77.1 & \textbf{66.1} & 69.4 & \textbf{64.3} & \textbf{79.7} \\
\textbf{Flan-UL2 Zero-shot} & 32.0 & 34.6 & 13.6 & 18.7 & 60.8 & 69.4 & 52.4 & 59.5 & 63.0 & \textbf{71.1} & 57.0 & 68.3 \\
\textbf{GPT-4 Zero-shot} & 46.2 & 51.5 & 45.4 & \textbf{65.1} & \textbf{66.4} & \textbf{74.3} & \textbf{67.3} & \textbf{82.5} & - & - & - & - \\
\textbf{GPT-4 Few-shot} & \textbf{48.3} & \textbf{54.3} & \textbf{45.5} & 61.3 & 63.3 & 73.6 & 61.2 & 73.5 & - & - & - & - \\
\bottomrule

\end{tabular}
}
\label{tab:main_results}
\vspace{-10pt}
\end{table*}

\noindent
\textbf{Training details, hyperparameters and implementations.}\,
We use a preceding window of 3 and a proceeding window of 3 for all methods, including supervised ones and prompting methods, thus the input to language model includes up to 7 utterances except for the ones at the start or end of the dialogue. 
We additionally include an ablation study on using no proceeding utterances and find it mildly hurts performance (\cref{subsec:ablation}). 
For supervised methods, our implementation is based on a mixture of PyTorch and JAX. 
For prior methods in~\cref{subsec:baselines}, we adapt from their official codebases respectively.
We use official GPT-4 API (\texttt{GPT-4-0613})~\footnote{https://platform.openai.com/docs/guides/gpt/chat-completions-api}.
More training details is referred to~\cref{appendix:training_details}.

\section{Result and Analysis}
\label{sec:results}

\subsection{Main Results}

We present the main results in Table~\ref{tab:main_results}.
\textbf{We notice that prompting methods---although require no or less labeled training instances, have lower performance compared to supervised methods}.
For example, on EMH dataset, GPT-4 Zero-shot only achieves 45.4\% Macro F1 and 65.1\% accuracy, while GPT-4 Few-shot achieves 45.5\% and 61.3\%.
On the other hand, encoder models still perform competitively, outperforming prompting methods by a large margin and the improvement is consistent across all datasets.
For example, on EMH dataset, the best encoder model Flan-T5-Enc-XL achieves 76.6\% Macro F1 and 88.4\% accuracy.
This observation is consistent with recent works on designing challenging datasets for LLMs~\citep{ravichander-etal-2022-condaqa} and testing their understanding capabilities for implicatures conveyed in communications~\citep{ruis2022large}.
From another perspective, this also indicates the challenging nature of empathy understanding task---LLMs still struggle to understand empathy despite being provided with the exact definition in prompts, while supervised fine-tuning still outperforms prompting methods with sufficient training samples.

Interestingly, GPT-4 Zero-shot outperforms Few-shot in terms of accuracy on both EMH and ESConv dataset, which contradicts prior works' observation that adding few-shot examplers help improve performance~\citep{kojima2022large,wei2022chain,min-etal-2022-rethinking}.
We hypothesize that our task---dialogue understanding is different from tasks evaluated by prior works in the following aspects:
(1) empathy dialogue understanding is relatively challenging for the language models.
Here challenging means it requires background knowledge and expertise which might not be prevalent in the language model's pretraining corpus.
(2) dialogue understanding leads to long input length to the language model.
As suggested by recent works~\citep[][\interalia]{liu-etal-2021-towards}, processing long sequences is still challenging for LLMs.

\begin{table*}
\begin{minipage}[t]{.49\linewidth}
\caption{Ablation study on the effect of loss function. The results are averaged over 20 predictive tasks on Empeval. We highlight the best number in each column.\label{tab:ablation_loss}}
\centering
{\setlength{\extrarowheight}{1.5pt}
\begin{adjustbox}{max width=\linewidth}
\begin{tabular}{lrrrr}
\toprule
\begin{tabular}[c]{@{}l@{}l} \textbf{Loss Function} \\ \, \\ \end{tabular} 
& \begin{tabular}[c]{@{}l@{}l} \textbf{Macro} \\ \textbf{Pre.} \\ \end{tabular} 
& \begin{tabular}[c]{@{}l@{}l} \textbf{Macro} \\ \textbf{Rec.} \\ \end{tabular} 
& \begin{tabular}[c]{@{}l@{}l} \textbf{Macro} \\ \textbf{F1} \\ \end{tabular} 
& \begin{tabular}[c]{@{}l@{}l} \textbf{ACC}. \\ \, \\ \end{tabular} \\  \midrule
\textbf{Cross Entropy Loss} & $\textbf{80.5}$ & 75.8 & 77.2 & 92.1 \\
\textbf{LDAM Loss} & 79.8 & 76.8 & 77.5 & $\textbf{92.3}$ \\
\textbf{Focal Loss} & 79.5 & $\textbf{77.3}$ & $\textbf{78.0}$ & 92.1 \\
\bottomrule
\end{tabular}
\end{adjustbox}
}
\end{minipage}
\hfill
\begin{minipage}[t]{.49\linewidth}
\caption{Ablation study on model sizes. The results are based on focal loss, averaged over 20 predictive tasks on Empeval. We highlight the best number in each column.\label{tab:ablation_model_size}}
\centering
{\setlength{\extrarowheight}{1.5pt}
\begin{adjustbox}{max width=\linewidth}
\begin{tabular}{lrrrr}
\toprule
\begin{tabular}[c]{@{}l@{}l} \textbf{Backbone LM} \\ \, \\ \end{tabular} 
& \begin{tabular}[c]{@{}l@{}l} \textbf{Macro} \\ \textbf{Precision} \\ \end{tabular} 
& \begin{tabular}[c]{@{}l@{}l} \textbf{Macro} \\ \textbf{Recall} \\ \end{tabular} 
& \begin{tabular}[c]{@{}l@{}l} \textbf{Macro} \\ \textbf{F1} \\ \end{tabular} 
& \begin{tabular}[c]{@{}l@{}l} \textbf{ACC}. \\ \, \\ \end{tabular} \\  \midrule
\textbf{Flan-T5-Small (60M)} & 68.7 & 66.5 & 66.4 & 91.1 \\
\textbf{Flan-T5-Base (220M)} & 78.1 & 74.5 & 75.7 & 91.8 \\
\textbf{Flan-T5-Large (770M)} & 79.5 & $\textbf{77.3}$ & $\textbf{78.0}$ & $\mathbf{92.1}$ \\
\textbf{Flan-T5-XL (3B)} & $\textbf{80.0}$ & 76.5 & 77.5 & 91.7 \\
\bottomrule
\end{tabular}
\end{adjustbox}
}
\end{minipage}
\end{table*}
\begin{table*}
\begin{minipage}[t]{.49\textwidth}
\caption{Ablation study on the effect of adding proceeding context. The results are based on focal loss, averaged over 20 predictive tasks on Empeval dataset. We highlight the best number in each column.\label{tab:ablation_context}}
\centering
{\setlength{\extrarowheight}{1.5pt}
\begin{adjustbox}{max width=\linewidth}
\begin{tabular}{lrrrr}
\toprule
\begin{tabular}[c]{@{}l@{}l} \textbf{Classifier Input} \\ \, \\ \end{tabular} 
& \begin{tabular}[c]{@{}l@{}l} \textbf{Macro} \\ \textbf{Pre.} \\ \end{tabular} 
& \begin{tabular}[c]{@{}l@{}l} \textbf{Macro} \\ \textbf{Rec.} \\ \end{tabular} 
& \begin{tabular}[c]{@{}l@{}l} \textbf{Macro} \\ \textbf{F1} \\ \end{tabular} 
& \begin{tabular}[c]{@{}l@{}l} \textbf{ACC}. \\ \, \\ \end{tabular} \\  \midrule
\textbf{w/o proceedings} & 79.2 & 74.8 & 75.6 & 91.6 \\
\textbf{w/ proceedings} & $\textbf{79.5}$ & $\textbf{77.3}$ & $\textbf{78.0}$ & $\textbf{92.1}$ \\
\bottomrule
\end{tabular}
\end{adjustbox}
}
\end{minipage}
\hfill
{\setlength{\extrarowheight}{1.5pt}
\begin{minipage}[t]{.49\textwidth}
\caption{Ablation study on the effect of adding natural language instructions. The results are based on focal loss, averaged over 20 predictive tasks on Empeval dataset. We highlight the best number in each column.\label{tab:ablation_instruction}}
\centering
\begin{adjustbox}{max width=\linewidth}
\begin{tabular}{lrrrr}
\toprule
\begin{tabular}[c]{@{}l@{}l} \textbf{Instruction Type} \\ \, \\ \end{tabular} 
& \begin{tabular}[c]{@{}l@{}l} \textbf{Macro} \\ \textbf{Precision} \\ \end{tabular} 
& \begin{tabular}[c]{@{}l@{}l} \textbf{Macro} \\ \textbf{Recall} \\ \end{tabular} 
& \begin{tabular}[c]{@{}l@{}l} \textbf{Macro} \\ \textbf{F1} \\ \end{tabular} 
& \begin{tabular}[c]{@{}l@{}l} \textbf{ACC}. \\ \, \\ \end{tabular} \\  \midrule
\textbf{w/o instructions} & 76.1 & 71.3 & 72.4 & 91.1 \\
\textbf{w/ instructions} & $\textbf{79.5}$ & $\textbf{77.3}$ & $\textbf{78.0}$ & $\textbf{92.1}$ \\
\bottomrule
\end{tabular}
\end{adjustbox}
\end{minipage}
}
\end{table*}

\noindent
\textbf{Instruction finetuned models achieve the best performance compared to encoder models and prompting methods, the improvement is consistent on all datasets.}
On 2 out of 3 datasets, Flan-T5-Large (770M) achieves better performance compared to the best encoder model, only worse than Flan-T5-Enc-XL (1.2B) on EMH dataset.
Further, Flan-T5-XL (3B) achieves the best performance compared to all other methods. 
We draw the conclusion that instruction-finetuned methods are better suited to be used for measuring conversational empathy.

\subsection{Ablation Studies}
\label{subsec:ablation}

Here we explore several ablations to understand how different design choices affect instruction finetuned encoder-decoder model's performance.
Unless otherwise explicitly mentioned, the results reported in this subsection are based on Flan-T5-Large (770M) on our internal Empeval dataset.

\noindent
\textbf{Choice of loss functions.}\,
Most real-world datasets on empathy dialogue understanding are of longtail distribution,
We examine whether different choices of loss functions can improve the model's performance.
From Table~\ref{tab:ablation_loss} we can see that LDAM loss and Focal loss indeed improve classification performance on less frequent classes---evidenced by improved Macro recall and Macro F1, while still being competitive in overall accuracy.
This finding also suggests the possibility of further improving performance by carefully choosing loss function and corresponding hyperparameters tuning.

\noindent
\textbf{Effect of model sizes.}\,
In Table~\ref{tab:ablation_model_size} we show a comparison between different sizes of instruction-finetuned encoder-decoder models.
We can notice that as the model size scales from Small (60M) to Large (770M), the performance gets a significant leap, then plateaus around Large.
In fact, the 3B model has slightly worse Macro F1 compared to 770M model (77.5\% vs 78\%).
We hypothesize 3B model's slight worse performance might be because it quickly overfits the dataset.

\noindent
\textbf{Effect of proceeding contexts.}\,
As mentioned earlier, for our main results in Table~\ref{tab:main_results} we use a context window of 3 preceding utterances and 3 proceeding utterances. 
In production systems, it is critical to monitor the dialogue state in real-time, where the proceeding utterances are not available.

From Table~\ref{tab:ablation_context} we notice removing proceeding contexts hurts performance.
The result is expected as the model uses less input information, and matches our motivation that empathy is a collaborative practice and context-dependent.

\noindent
\textbf{Effect of natural language instructions.}\,
The effect of natural language instructions is shown in Table~\ref{tab:ablation_instruction}. 
Adding instructions significantly improves classification performance on all metrics.
This comparison highlights the importance of natural language instructions in instruction finetuning.

\section{Conclusion and Future Work}
In this study, we introduce a comprehensive framework for assessing conversational empathy, focusing on both expressed empathetic intentions and perceived empathy.
We apply this framework to analyze our internal dataset and find that it effectively correlates expressed intentions, perceived empathy, and overall dialogue satisfaction.

This proposed conceptual evaluation framework requires subjective assessments from trained annotators, which can be expensive and non-trivial to collect.
To explore automated empathy measurement models, we rigorously compare various methods against human judgments. 
Our instruction finetuning technique achieves the highest classification performance, measured by F-score and accuracy compared to human-based judgments. 
While our discussion primarily centers on measuring empathy perception in human-human conversations, an important future direction is extending this analysis to human-machine interactions.

\section*{Limitations and Potential Risks}

This paper studies the empathy evaluation problem in the context of human-human communication. 
A natural extension is to apply our framework to dialogue applications, and the findings in this paper may be subject to change in human-AI dialogues. 
Due to limited bandwidth, we only experiment with one proprietary model---\texttt{GPT-4-0613}. Including more latest proprietary models such as GPT-4o, Gemini~\cite{team2023gemini}, and Claude will make the results more comprehensive.

To the best of our knowledge, this study does not involve potential risks.



\bibliography{anthology}

\appendix
\section{Additional Information on Conversation Analysis}
In Table~\ref{tab:relationship_intent_empathy_w_stderr}, we report the mean and standard error of the perceived empathy dimension, when an intent is present or not.

\begin{table*}[!h]
\centering
\caption{
Relationship between expressed intents and perceived empathy dimensions and user satisfaction (each column reports the mean and standard error of the perceived empathy dimension's ratings when an intent appears (True) or not (False)); $\diamondsuit$, $\dag$, and $\ddag$ indicate a statistical significance of ratings when the intent appears or not at $p<0.05$, $0.01$, and $0.001$, respectively.
}
\vspace{10pt}
\resizebox{\textwidth}{!}{
\begin{tabular}{l|ccc|ccc|ccc|ccc||ccc}

\hline

\textbf{Expressed Intent} & \multicolumn{3}{c|}{\textbf{Perceived Enthusiasm}} & \multicolumn{3}{c|}{\textbf{Perceived Understand}} & \multicolumn{3}{c|}{\textbf{Perceived Sympathy}} & \multicolumn{3}{c||}{\textbf{Perceived Helpfulness}} & \multicolumn{3}{c}{\textbf{Conversation Satisfaction}} \\ \cline{2-16}

 & Intent=True & Intent=False & & Intent=True & Intent=False & & Intent=True & Intent=False & & Intent=True & Intent=False & & Intent=True & Intent=False & \\ \hline

ask contact                      &   3.97 (0.01)   &   3.99 (0.01)  &        &   3.90 (0.02) &  3.96 (0.01) &   $\dag$  &   3.05 (0.01) &  3.08 (0.01) &    $\diamondsuit$  &   3.83 (0.02) &  3.87 (0.01) &    $\diamondsuit$    &   3.72 (0.03) &  3.74 (0.01) &       \\
ask details                      &   4.01 (0.01)   &   3.98 (0.01)  &        &   3.95 (0.02) &  3.95 (0.01) &       &   3.07 (0.01) &  3.08 (0.01) &       &   3.89 (0.02) &  3.86 (0.01) &         &   3.75 (0.03) &  3.74 (0.01) &       \\
ask confirm                      &   4.00 (0.02)   &   3.99 (0.01)  &        &   3.96 (0.02) &  3.95 (0.01) &       &   3.06 (0.02) &  3.07 (0.01) &       &   3.92 (0.03) &  3.86 (0.01) &    $\diamondsuit$    &   3.85 (0.04) &  3.73 (0.01) &   $\dag$  \\
aware problem                    &   4.03 (0.01)   &   3.97 (0.01)  &  $\ddag$   &   4.03 (0.01) &  3.91 (0.01) &  $\ddag$  &   3.12 (0.01) &  3.06 (0.01) &  $\ddag$  &   3.92 (0.01) &  3.84 (0.01) &  $\ddag$    &   3.85 (0.02) &  3.69 (0.01) &  $\ddag$  \\
describe problem                 &   4.04 (0.01)   &   3.98 (0.01)  &  $\ddag$   &   4.02 (0.01) &  3.94 (0.01) &  $\ddag$  &   3.15 (0.02) &  3.06 (0.01) &  $\ddag$  &   3.93 (0.01) &  3.85 (0.01) &  $\ddag$    &   3.88 (0.02) &  3.72 (0.01) &  $\ddag$  \\
express sympathy                 &   4.07 (0.02)   &   3.98 (0.01)  &  $\ddag$   &   4.01 (0.02) &  3.94 (0.01) &   $\dag$  &   3.48 (0.04) &  3.04 (0.00) &  $\ddag$  &   3.87 (0.03) &  3.86 (0.01) &         &   3.82 (0.04) &  3.74 (0.01) &       \\
express reassurance              &   4.06 (0.01)   &   3.97 (0.01)  &  $\ddag$   &   4.01 (0.01) &  3.94 (0.01) &  $\ddag$  &   3.17 (0.02) &  3.05 (0.01) &  $\ddag$  &   3.93 (0.02) &  3.85 (0.01) &  $\ddag$    &   3.84 (0.02) &  3.72 (0.01) &  $\ddag$  \\
express apology                  &   3.97 (0.02)   &   3.99 (0.01)  &        &   3.96 (0.02) &  3.95 (0.01) &       &   3.12 (0.02) &  3.07 (0.01) &    $\diamondsuit$  &   3.84 (0.02) &  3.87 (0.01) &         &   3.69 (0.03) &  3.75 (0.01) &       \\
answer question                  &   4.08 (0.02)   &   3.98 (0.01)  &  $\ddag$   &   4.05 (0.03) &  3.94 (0.01) &  $\ddag$  &   3.08 (0.03) &  3.07 (0.01) &       &   4.02 (0.03) &  3.85 (0.01) &  $\ddag$    &   3.92 (0.04) &  3.73 (0.01) &  $\ddag$  \\
clarify                          &   4.06 (0.05)   &   3.99 (0.01)  &        &   4.11 (0.05) &  3.95 (0.01) &  $\ddag$  &   3.14 (0.06) &  3.07 (0.01) &       &   4.06 (0.05) &  3.86 (0.01) &  $\ddag$    &   3.99 (0.07) &  3.74 (0.01) &   $\dag$  \\
explain                          &   4.09 (0.03)   &   3.99 (0.01)  &   $\dag$   &   4.08 (0.03) &  3.95 (0.01) &  $\ddag$  &   3.03 (0.03) &  3.08 (0.01) &       &   4.07 (0.03) &  3.86 (0.01) &  $\ddag$    &   4.00 (0.05) &  3.73 (0.01) &  $\ddag$  \\
excuse                           &   3.96 (0.04)   &   3.99 (0.01)  &        &   3.97 (0.04) &  3.95 (0.01) &       &   3.08 (0.04) &  3.07 (0.01) &       &   3.78 (0.06) &  3.87 (0.01) &         &   3.56 (0.08) &  3.75 (0.01) &    $\diamondsuit$  \\
inform action                    &   4.06 (0.02)   &   3.98 (0.01)  &  $\ddag$   &   4.03 (0.02) &  3.94 (0.01) &  $\ddag$  &   3.08 (0.02) &  3.07 (0.01) &       &   3.98 (0.02) &  3.85 (0.01) &  $\ddag$    &   3.90 (0.03) &  3.72 (0.01) &  $\ddag$  \\
instruct action                  &   4.02 (0.02)   &   3.98 (0.01)  &   $\diamondsuit$    &   3.97 (0.02) &  3.95 (0.01) &       &   3.08 (0.02) &  3.07 (0.01) &       &   3.91 (0.02) &  3.86 (0.01) &    $\diamondsuit$    &   3.75 (0.03) &  3.74 (0.01) &       \\
tentative solution               &   4.10 (0.04)   &   3.98 (0.01)  &   $\dag$   &   4.06 (0.04) &  3.95 (0.01) &   $\dag$  &   3.16 (0.04) &  3.07 (0.01) &    $\diamondsuit$  &   4.00 (0.04) &  3.86 (0.01) &   $\dag$    &   3.87 (0.06) &  3.74 (0.01) &    $\diamondsuit$  \\
contact other                    &   3.95 (0.03)   &   3.99 (0.01)  &        &   3.88 (0.03) &  3.95 (0.01) &    $\diamondsuit$  &   3.05 (0.02) &  3.08 (0.01) &       &   3.74 (0.04) &  3.87 (0.01) &   $\dag$    &   3.66 (0.05) &  3.75 (0.01) &       \\
\hline
\end{tabular}
}
\label{tab:relationship_intent_empathy_w_stderr}
\end{table*}

\section{Annotation of Empeval}
\label{appendix:detailed_annotation}

We show the annotation questions in Table~\ref{tab:annotation_questions_empeval_p1} and Table~\ref{tab:annotation_questions_empeval_p2}.
The collected dialogues are used for academic purposes with agreements from the users.
Our annotation process is approved by our internal ethics review board. 

\begin{table*}[!h]
\centering
\caption{Annotation questions for Empeval dataset --- expressed communicative intents.}
\vspace{0pt}
\resizebox{\textwidth}{!}{
\begin{tabular}{ll}
\toprule
Ask Contact
& \begin{tabular}[c]{@{}p{1\linewidth}@{}}  
Did the highlighted messages ask the customer for contact information (e.g., name, email, phone number, ID)?\\
\textbullet \, Yes\\
\textbullet \, No
\end{tabular}\\
\midrule
Ask Details
& \begin{tabular}[c]{@{}p{1\linewidth}@{}}  
Did the highlighted messages ask the customer to provide problem details (excluding contact information)?\\
\textbullet \, Yes\\
\textbullet \, No
\end{tabular}\\
\midrule
Ask Confirmation
& \begin{tabular}[c]{@{}p{1\linewidth}@{}}  
Did the highlighted messages ask the customer to confirm or clarify problem details (excluding contact information)?\\
\textbullet \, Yes\\
\textbullet \, No
\end{tabular}\\
\midrule
Aware Problem
& \begin{tabular}[c]{@{}p{1\linewidth}@{}}  
Did the highlighted messages express that the agent was aware of the customer’s problem?\\
\textbullet \, Yes\\
\textbullet \, No
\end{tabular}\\
\midrule
Describe Problem
& \begin{tabular}[c]{@{}p{1\linewidth}@{}}  
Did the highlighted messages describe, repeat, or paraphrase the customer’s problem? (e.g., “I understand you have problems with refund.”)\\
\textbullet \, Yes\\
\textbullet \, No
\end{tabular}\\
\midrule
Express Sympathy
& \begin{tabular}[c]{@{}p{1\linewidth}@{}}  
Did the highlighted messages express sympathy to the customer? (e.g., “I am sorry that your refund was denied. I know how frustrating this feels.”)\\
\textbullet \, Yes\\
\textbullet \, No
\end{tabular}\\
\midrule
Express Reassurance
& \begin{tabular}[c]{@{}p{1\linewidth}@{}}  
Did the highlighted messages express reassurance and support to the customer? (e.g., “Please don’t worry. I will try my best to help you.”)\\
\textbullet \, Yes\\
\textbullet \, No
\end{tabular}\\
\midrule
Express Apology
& \begin{tabular}[c]{@{}p{1\linewidth}@{}}  
Did the highlighted messages express apology to the customer? (e.g., “I really apologize for the inconveniences.”)\\
\textbullet \, Yes\\
\textbullet \, No
\end{tabular}\\
\midrule
Answer Question
& \begin{tabular}[c]{@{}p{1\linewidth}@{}}  
Did the highlighted messages answer the customer’s questions?\\
\textbullet \, Yes\\
\textbullet \, No
\end{tabular}\\
\midrule
Clarify
& \begin{tabular}[c]{@{}p{1\linewidth}@{}}  
Did the highlighted messages clarify the customer’s confusions or misunderstandings?\\
\textbullet \, Yes\\
\textbullet \, No
\end{tabular}\\
\midrule
Explain
& \begin{tabular}[c]{@{}p{1\linewidth}@{}}  
Did the highlighted messages explain why the problem happened?\\
\textbullet \, Yes\\
\textbullet \, No
\end{tabular}\\
\midrule
Excuse
& \begin{tabular}[c]{@{}p{1\linewidth}@{}}  
Did the highlighted messages explain why the problem could not be solved?\\
\textbullet \, Yes\\
\textbullet \, No
\end{tabular}\\
\midrule
Inform Action
& \begin{tabular}[c]{@{}p{1\linewidth}@{}}  
Did the highlighted messages inform the customer that the agent had taken actions to solve the problem?\\
\textbullet \, Yes\\
\textbullet \, No
\end{tabular}\\
\midrule
Instruct Action
& \begin{tabular}[c]{@{}p{1\linewidth}@{}}  
Did the highlighted messages provide specific instructions that the customer could take to solve the problem?\\
\textbullet \, Yes\\
\textbullet \, No
\end{tabular}\\
\midrule
Tentative Solution
& \begin{tabular}[c]{@{}p{1\linewidth}@{}}  
Did the highlighted messages offer a tentative solution to the customer?\\
\textbullet \, Yes\\
\textbullet \, No
\end{tabular}\\
\midrule
Contact Others
& \begin{tabular}[c]{@{}p{1\linewidth}@{}}  
Did the highlighted messages ask the customer to contact someone else for the issue?\\
\textbullet \, Yes\\
\textbullet \, No
\end{tabular}\\
\bottomrule
\end{tabular}
}
\label{tab:annotation_questions_empeval_p1}
\vspace{0pt}
\end{table*}

\begin{table*}[h]
\centering
\caption{Annotation questions for Empeval dataset --- perceived empathy dimensions.}
\vspace{0pt}
\resizebox{\textwidth}{!}{
\begin{tabular}{ll}
\toprule
Perceived Enthusiasm
& \begin{tabular}[c]{@{}p{1\linewidth}@{}}  
 If I were the customer, I would believe the agent was enthusiastic to help me after receiving the highlighted messages.\\
\textbullet \, Strongly disagree\\
\textbullet \, Disagree\\
\textbullet \, Neither\\
\textbullet \, Agree\\
\textbullet \, Strongly agree
\end{tabular}\\
\midrule
Perceived Helpfulness
& \begin{tabular}[c]{@{}p{1\linewidth}@{}}  
If I were the customer, I would believe the agent helped me effectively after receiving the highlighted messages.\\
\textbullet \, Strongly disagree\\
\textbullet \, Disagree\\
\textbullet \, Neither\\
\textbullet \, Agree\\
\textbullet \, Strongly agree
\end{tabular} \\ 
\midrule
Perceived Sympathy
& \begin{tabular}[c]{@{}p{1\linewidth}@{}}  
If I were the customer, I would believe the agent sympathized with my feelings properly after receiving the highlighted messages.\\
\textbullet \, Strongly disagree\\
\textbullet \, Disagree\\
\textbullet \, Neither\\
\textbullet \, Agree\\
\textbullet \, Strongly agree
\end{tabular} \\ 
\midrule
Perceived Understanding
& \begin{tabular}[c]{@{}p{1\linewidth}@{}}  
If I were the customer, I would believe the agent understood my problem or request accurately after receiving the highlighted messages.\\
\textbullet \, Strongly disagree\\
\textbullet \, Disagree\\
\textbullet \, Neither\\
\textbullet \, Agree\\
\textbullet \, Strongly agree
\end{tabular} \\ 
\bottomrule
\end{tabular}
}
\label{tab:annotation_questions_empeval_p2}
\vspace{0pt}
\end{table*}

\section{More Details of Datasets}
\label{appendix:datasets}
EMH dataset~\cite{sharma-etal-2020-computational} is a multi-label classification dataset, where each dialogue session is annotated on three expressed empathy dimensions: \emph{emotional reactions}, \emph{interpretations} and \emph{explorations} at three different levels: \emph{no communicative intent}, \emph{weak communicative intent} and \emph{strong communicative intent}.

ESConv~\cite{liu-etal-2021-towards} is annotated with 3 coarse-grained mental support stages: \emph{explorations}, \emph{comforting} and \emph{action}, and seven fine-grained mental support strategies: \emph{question}, \emph{restatement or paraphrasing}, \emph{reflection of feelings}, \emph{self-disclosure}, \emph{affirmation and reassurance}, \emph{providing suggestions} and \emph{information}.

We present the statistics of dataset in Table~\ref{tab:dataset_statistics}. Both public datasets' licenses allow for usage for academic purposes.

\begin{table*}[!h]
\centering
\caption{
Dataset statistics, as ESConv and Empeval have binary labels, we mark them as strong intent and no intent.
}
\vspace{0pt}
\resizebox{0.8\linewidth}{!}{
\begin{tabular}{lccc}
\toprule
& \# No Intent & \# Weak Intent & \# Strong Intent
\\ 
\rowcolor{lightgray}
\multicolumn{4}{l}{\emph{\textbf{Empathy Mental Health (EMH)}}}\\
Emotional Reactions & 2,037 & 895 & 152 \\
Interpretations & 1,626 & 114 & 1,344 \\
Explorations & 2,604 & 104 & 376 \\
\rowcolor{lightgray}
\multicolumn{4}{l}{\emph{\textbf{ESConv}}}\\
\hline
Question & 14,557 & N.A. & 3,799 \\
Affirmation \& Reassurance & 15,533 & N.A. & 2,823 \\
Providing Suggestions & 15,408 & N.A. & 2,948 \\
Information & 17,142 & N.A. & 1,214 \\
Self-disclosure & 16,645 & N.A. & 1,711 \\
Restatement or Paraphrase & 17,267 & N.A. & 1,089 \\
Reflection of Feelings & 16,922 & N.A. & 1,434 \\
\rowcolor{lightgray}
\multicolumn{4}{l}{\emph{\textbf{Empeval}}}\\
\hline
Ask Contact & 1,552 & N.A. & 281 \\
Ask Details & 1,553 & N.A. & 280 \\
Ask Confirmation & 1,692 & N.A. & 141 \\
Aware Problem & 1,038 & N.A. & 795 \\
Describe Problem & 1,505 & N.A. & 328 \\
Express Sympathy & 1,658 & N.A. & 175 \\
Express Reassurance & 1,447 & N.A. & 386 \\
Express Apology & 1,588 & N.A. & 245 \\
Answer Question & 1,654 & N.A. & 179 \\
Clarify & 1,773 & N.A. & 60 \\
Explain & 1,771 & N.A. & 62 \\
Excuse & 1,761 & N.A. & 72 \\
Inform Action & 1,576 & N.A. & 257 \\
Instruct Action & 1,553 & N.A. & 280 \\
Tentative Solution & 1,736 & N.A. & 97 \\
Contact Others & 1,678 & N.A. & 155 \\ \midrule
Perceived Enthusiasm & 121 & N.A. & 1,712 \\
Perceived Helpfulness & 208 & N.A. & 1,625 \\
Perceived Sympathy & 1,568 & N.A. & 265 \\
Perceived Understanding & 400 & N.A. & 1,433 \\
\bottomrule
\end{tabular}
}
\label{tab:dataset_statistics}
\vspace{0pt}
\end{table*}

\section{Additional Training Details}
\label{appendix:training_details}
For prior methods, we adapt their official implementations for our predictive tasks.
For the four encoder models (BERT-Large, RoBERTa-Large, Flan-T5-Enc-Large, Flan-T5-Enc-XL), our implementation is based on PyTorch. 
For two instruction-finetuned encoder-decoder models, i.e. Flan-T5-Large and Flan-T5-XL, we use T5X framework~\cite{roberts2022t5x} based on JAX.

For all models, we use AdamW optimizer, a smaller learning rate of 5e-6 and train in total 30 epochs. Additional early stopping is used to avoid overfitting.

For instruction finetuning encoder-decoder models, we use the same zero-shot template as we use for prompting methods, and train the model to generate the corresponding verbalizers. 
In testing, we use the rank classification strategy, i.e. we choose the label with the highest probability assigned to the corresponding verbalizer token(s).

\section{Prompt Template}
\label{appendix:prompt_template}
An author of this paper manually crafted a set of prompt templates and validate the performance on a held-out validation set to pick the highest-performing ones.
We show the prompt template we use for Empathy Mental Health dataset in Table~\ref{tab:template_emh}, ESConv dataset in Table~\ref{tab:template_esconv} and Empeval in Table~\ref{tab:template_empeval}.
\begin{table*}[h]
\centering
\caption{Prompt template for Empathy Mental Health (EMH) dataset.}
\vspace{0pt}
\resizebox{\textwidth}{!}{
\begin{tabular}{ll}
\toprule
Emotional Reactions & 
\begin{tabular}[c]{@{}p{1\linewidth}@{}}  
You are a crowdsourcing annotator. Now read the following definition and a corresponding dialogue between an emotional support seeker and an emotional supporter provider, then answer the corresponding question.\\
Definition: emotional reactions is a common strategy in online mental support. It refers to the supporter expressing emotions such as warmth, compassion, and concern about what was experienced by the help seeker after listening to the help seeker's utterance in a dialogue.\\
A week communication address those emotions without explicit reference, e.g. 'Everything will be fine', while strong communication specifies the emotion, e.g. 'I feel really sorry for you.'\\
\{Dialogue\} \\
Question: by saying \{utterance\}, what is the extent of emotional reactions expressed by the supporter? Respond with strong, weak or no communication. \\
\end{tabular}\\
\midrule
Interpretations & 
\begin{tabular}[c]{@{}p{1\linewidth}@{}}  
You are a crowdsourcing annotator. Now read the following definition and a corresponding dialogue between an emotional support seeker and an emotional supporter provider, then answer the corresponding question.\\ 
Definition: interpretations is a common strategy in online mental support. It refers to the supporter communicating an understanding of feelings and experiences inferred from the help seeker's utterance in the dialogue. A weak communication of interpretations contains a mention of the understanding, e.g. 'I understand how you feel', while a strong communication specifies the inferred feeling or experience, e.g. 'This must be terrifying' or communicates understanding through describing similar experiences, e.g. 'I also have the same experience'.\\
\{Dialogue\} \\
Question: by saying \{utterance\}, what is the extent of interpretations expressed by the supporter? Respond with strong, weak or no communication. \\
\end{tabular}\\
\midrule
Interpretations & 
\begin{tabular}[c]{@{}p{1\linewidth}@{}}  
You are a crowdsourcing annotator. Now read the following definition and a corresponding dialogue between an emotional support seeker and an emotional supporter provider, then answer the corresponding question.\\
Definition: explorations is a common strategy in online mental support. It refers to the supporter tries to improve the understanding of the help seeker by exploring the feelings and experiences not explicitly stated in the previous context of the dialogue. A weak communication of explorations is usually to ask a generic question, e.g. 'What happened', while a strong communication of explorations is to explicitly mention the help seeker's experiences and feelings which the supporter wants to explore, e.g. 'Are you feeling lonely right now?'.\\
\{Dialogue\} \\
Question: by saying \{utterance\}, what is the extent of explorations expressed by the supporter? Respond with strong, weak or no communication. \\
\end{tabular}\\
\bottomrule
\end{tabular}
}
\label{tab:template_emh}
\vspace{0pt}
\end{table*}

\begin{table*}[h]
\centering
\caption{Prompt template for ESConv dataset.}
\vspace{0pt}
\resizebox{\textwidth}{!}{
\begin{tabular}{ll}
\toprule
Question 
& \begin{tabular}[c]{@{}p{1\linewidth}@{}}  
You are a crowdsourcing annotator. Now read the following dialogue between a customer and a therapist, then answering a corresponding question.\\
\{Dialogue\}\\
Question: by saying \{utterance\}, is the therapist asking the customer a question? Respond with yes or no.
\end{tabular}\\
\midrule
Affirmation \& Reassurance 
& \begin{tabular}[c]{@{}p{1\linewidth}@{}}  
You are a crowdsourcing annotator. Now read the following dialogue between a customer and a therapist, then answering a corresponding question.\\
\{Dialogue\}\\
Question: by saying \{utterance\}, is the therapist showing affirmation or encouraging the customer? Respond with yes or no.
\end{tabular}\\
\midrule
Providing Suggestions
& \begin{tabular}[c]{@{}p{1\linewidth}@{}}  
You are a crowdsourcing annotator. Now read the following dialogue between a customer and a therapist, then answering a corresponding question.\\
\{Dialogue\}\\
Question: by saying \{utterance\}, is the therapist providing suggestions to the customer? Respond with yes or no.
\end{tabular}\\
\midrule
Information
& \begin{tabular}[c]{@{}p{1\linewidth}@{}}  
You are a crowdsourcing annotator. Now read the following dialogue between a customer and a therapist, then answering a corresponding question.\\
\{Dialogue\}\\
Question: by saying \{utterance\}, is the therapist providing the customer with additional background information or facts? Respond with yes or no.
\end{tabular}\\
\midrule
Self disclosure
& \begin{tabular}[c]{@{}p{1\linewidth}@{}}  
You are a crowdsourcing annotator. Now read the following dialogue between a customer and a therapist, then answering a corresponding question.\\
\{Dialogue\}\\
Question: by saying \{utterance\}, is the therapist saying they share the same feeling or experience as the customer? Respond with yes or no.
\end{tabular}\\
\midrule
Restatement or Paraphrasing
& \begin{tabular}[c]{@{}p{1\linewidth}@{}}  
You are a crowdsourcing annotator. Now read the following dialogue between a customer and a therapist, then answering a corresponding question.\\
\{Dialogue\}\\
Question: by saying \{utterance\}, is the therapist trying to paraphrase or clarify what the customer just talked about? Respond with yes or no.
\end{tabular}\\
\midrule
Reflection of Feelings
& \begin{tabular}[c]{@{}p{1\linewidth}@{}}  
You are a crowdsourcing annotator. Now read the following dialogue between a customer and a therapist, then answering a corresponding question.\\
\{Dialogue\}\\
Question: by saying \{utterance\}, is the therapist trying to say they can understand what the customer is currently feeling or experiencing? Respond with yes or no.
\end{tabular}\\
\bottomrule
\end{tabular}
}
\label{tab:template_esconv}
\vspace{0pt}
\end{table*}

\begin{table*}[h]
\centering
\caption{Prompt template for Empeval dataset, part 1.}
\vspace{0pt}
\resizebox{\textwidth}{!}{
\begin{tabular}{ll}
\toprule
Ask Contact
& \begin{tabular}[c]{@{}p{1\linewidth}@{}}  
You are a crowdsourcing annotator. Now read the following dialogue between a customer and an agent, then answering a corresponding question.\\
\{Dialogue\}\\
Question: by saying \{utterance\}, is the agent trying to ask the contact information of the customer? Respond with yes or no.
\end{tabular}\\
\midrule
Ask Details
& \begin{tabular}[c]{@{}p{1\linewidth}@{}}  
You are a crowdsourcing annotator. Now read the following dialogue between a customer and an agent, then answering a corresponding question.\\
\{Dialogue\}\\
Question: by saying \{utterance\}, is the agent trying to ask the customer about the problem for more details? Respond with yes or no.
\end{tabular}\\
\midrule
Ask Confirmation
& \begin{tabular}[c]{@{}p{1\linewidth}@{}}  
You are a crowdsourcing annotator. Now read the following dialogue between a customer and an agent, then answering a corresponding question.\\
\{Dialogue\}\\
Question: by saying \{utterance\}, is the agent trying to ask the customer to confirm or clarify the details of the problem? Respond with yes or no.
\end{tabular}\\
\midrule
Aware Problem
& \begin{tabular}[c]{@{}p{1\linewidth}@{}}  
You are a crowdsourcing annotator. Now read the following dialogue between a customer and an agent, then answering a corresponding question.\\
\{Dialogue\}\\
Question: by saying \{utterance\}, is the agent trying to say they are aware of the problem? Respond with yes or no.
\end{tabular}\\
\midrule
Describe Problem
& \begin{tabular}[c]{@{}p{1\linewidth}@{}}  
You are a crowdsourcing annotator. Now read the following dialogue between a customer and an agent, then answering a corresponding question.\\
\{Dialogue\}\\
Question: by saying \{utterance\}, is the agent trying to describe, repeat of paraphrase the customer's question? Respond with yes or no.
\end{tabular}\\
\midrule
Express Sympathy
& \begin{tabular}[c]{@{}p{1\linewidth}@{}}  
You are a crowdsourcing annotator. Now read the following dialogue between a customer and an agent, then answering a corresponding question.\\
\{Dialogue\}\\
Question: by saying \{utterance\}, is the agent trying to express sympathy to the customer or to comfort the customer? Respond with yes or no.
\end{tabular}\\
\midrule
Express Reassurance
& \begin{tabular}[c]{@{}p{1\linewidth}@{}}  
You are a crowdsourcing annotator. Now read the following dialogue between a customer and an agent, then answering a corresponding question.\\
\{Dialogue\}\\
Question: by saying \{utterance\}, is the agent trying to express reassurance and support to the customer? Respond with yes or no.
\end{tabular} \\ 
\midrule
Express Apology
& \begin{tabular}[c]{@{}p{1\linewidth}@{}}  
You are a crowdsourcing annotator. Now read the following dialogue between a customer and an agent, then answering a corresponding question.\\
\{Dialogue\}\\
Question: by saying \{utterance\}, is the agent trying to apologize for what caused the customer's unpleasant experience? Respond with yes or no.
\end{tabular} \\ 
\midrule
Answer Question
& \begin{tabular}[c]{@{}p{1\linewidth}@{}}  
You are a crowdsourcing annotator. Now read the following dialogue between a customer and an agent, then answering a corresponding question.\\
\{Dialogue\}\\
Question: by saying \{utterance\}, is the agent trying to answer the questions raised by the customer? Respond with yes or no.
\end{tabular} \\ 
\midrule
Clarify
& \begin{tabular}[c]{@{}p{1\linewidth}@{}}  
You are a crowdsourcing annotator. Now read the following dialogue between a customer and an agent, then answering a corresponding question.\\
\{Dialogue\}\\
Question: by saying \{utterance\}, is the agent trying to clarify confusions or misunderstandings in the preceding context of the dialogue? Respond with yes or no.
\end{tabular} \\ 
\bottomrule
\end{tabular}
}
\label{tab:template_empeval}
\vspace{0pt}
\end{table*}

\begin{table*}[h]
\centering
\caption{Prompt template for Empeval dataset, part 2.}
\vspace{0pt}
\resizebox{\textwidth}{!}{
\begin{tabular}{ll}
\toprule
Explain
& \begin{tabular}[c]{@{}p{1\linewidth}@{}}  
You are a crowdsourcing annotator. Now read the following dialogue between a customer and an agent, then answering a corresponding question.\\
\{Dialogue\}\\
Question: by saying \{utterance\}, is the agent trying to explain why the problem described by the customer happened? Respond with yes or no.
\end{tabular} \\ 
\midrule
Excuse
& \begin{tabular}[c]{@{}p{1\linewidth}@{}}  
You are a crowdsourcing annotator. Now read the following dialogue between a customer and an agent, then answering a corresponding question.\\
\{Dialogue\}\\
Question: by saying \{utterance\}, is the agent trying to explain why the problem described by the customer cannot be solved? Respond with yes or no.
\end{tabular} \\ 
\midrule
Inform Action
& \begin{tabular}[c]{@{}p{1\linewidth}@{}}  
You are a crowdsourcing annotator. Now read the following dialogue between a customer and an agent, then answering a corresponding question.\\
\{Dialogue\}\\
Question: by saying \{utterance\}, is the agent trying to say that actions has been taken to solve the problem? Respond with yes or no.
\end{tabular} \\ 
\midrule
Instruct Action
& \begin{tabular}[c]{@{}p{1\linewidth}@{}}  
You are a crowdsourcing annotator. Now read the following dialogue between a customer and an agent, then answering a corresponding question.\\
\{Dialogue\}\\
Question: by saying \{utterance\}, is the agent trying to provide specific instructions to the customer? Respond with yes or no.
\end{tabular} \\ 
\midrule
Tentative Solution
& \begin{tabular}[c]{@{}p{1\linewidth}@{}}  
You are a crowdsourcing annotator. Now read the following dialogue between a customer and an agent, then answering a corresponding question.\\
\{Dialogue\}\\
Question: by saying \{utterance\}, is the agent trying to provide the Customer with some temporary solutions? Respond with yes or no.
\end{tabular} \\ 
\midrule
Contact Others
& \begin{tabular}[c]{@{}p{1\linewidth}@{}}  
You are a crowdsourcing annotator. Now read the following dialogue between a customer and an agent, then answering a corresponding question.\\
\{Dialogue\}\\
Question: by saying \{utterance\}, is the agent trying to ask the customer to contact others for support or help? Respond with yes or no.
\end{tabular} \\ 
\midrule
Perceived Enthusiasm
& \begin{tabular}[c]{@{}p{1\linewidth}@{}}  
You are a crowdsourcing annotator. Now read the following dialogue between a customer and an agent, then answering a corresponding question.\\
\{Dialogue\}\\
Question: Given the above dialogue, can we say the agent is enthusiastic and eager to help the customer? Respond with yes or no.
\end{tabular} \\ 
\midrule
Perceived Helpfulness
& \begin{tabular}[c]{@{}p{1\linewidth}@{}}  
You are a crowdsourcing annotator. Now read the following dialogue between a customer and an agent, then answering a corresponding question.\\
\{Dialogue\}\\
Question: Given the above dialogue, can we say the agent has effectively helped the customer to solve the problem? Respond with yes or no.
\end{tabular} \\ 
\midrule
Perceived Sympathy
& \begin{tabular}[c]{@{}p{1\linewidth}@{}}  
You are a crowdsourcing annotator. Now read the following dialogue between a customer and an agent, then answering a corresponding question.\\
\{Dialogue\}\\
Question: Given the above dialogue, can we say the agent showed sympathy during the communications with the customer? Respond with yes or no.
\end{tabular} \\ 
\midrule
Perceived Understanding
& \begin{tabular}[c]{@{}p{1\linewidth}@{}}  
You are a crowdsourcing annotator. Now read the following dialogue between a customer and an agent, then answering a corresponding question.\\
\{Dialogue\}\\
Question: Given the above dialogue, can we say the agent has fully understood the problem encountered or raised by the customer? Respond with yes or no.
\end{tabular} \\ 
\bottomrule
\end{tabular}
}
\label{tab:template_empeval_1}
\vspace{0pt}
\end{table*}

\section{Full Results}
\label{subsec:full_results}
We report the full results for Empathy Mental Health dataset in Table~\ref{tab:emh_full_results} and ESConv dataset in Table~\ref{tab:esconv_full_results_p1} and Table~\ref{tab:esconv_full_results_p2}.

\begin{table*}[t]
\centering
\caption{Full results on Empathy Mental Health dataset.}
\vspace{0pt}
\resizebox{\textwidth}{!}{
\begin{tabular}{lrrrrrrrrrrrrr}
\toprule
\, & Precision & Recall & F1 & Precision & Recall & F1 & Precision & Recall & F1 & Macro-Precision & Macro-Recall & Macro-F1 & Accuracy \\
\midrule
Emotional Reactions & \multicolumn{3}{c}{No Intent} & \multicolumn{3}{c}{Weak Intent} & \multicolumn{3}{c}{Strong Intent} & \multicolumn{3}{c}{\,} \\

\rowcolor{lightgray}
\multicolumn{14}{l}{\textsc{Prior methods}} \\
\citet{sharma-etal-2020-computational}   & 0.824                & 0.878                & 0.850                & 0.715                & 0.643                & 0.677                & 0.815                & 0.688                & 0.746                & 0.785                & 0.736                & 0.758                & 0.794                \\
\citet{welivita2021large} & 0.835                & 0.829                & 0.832                & 0.638                & 0.603                & 0.620                & 0.500                & 0.719                & 0.590                & 0.658                & 0.717                & 0.681                & 0.792                \\
\citet{li-etal-2023-understanding}      & 0.854                & 0.862                & 0.858                & 0.709                & 0.724                & 0.716                & 0.917                & 0.688                & 0.786                & 0.827                & 0.758                & 0.787                & 0.808                \\
\rowcolor{lightgray}
\multicolumn{14}{l}{\textsc{Encoder Models}} \\
BERT-Large           & 0.869                & 0.875                & 0.872                & 0.731                & 0.739                & 0.735                & 0.889                & 0.750                & 0.814                & 0.830                & 0.788                & 0.807                & 0.825                \\
RoBERTa-Large        & 0.879                & 0.909                & 0.894                & 0.787                & 0.744                & 0.765                & 0.800                & 0.750                & 0.774                & 0.822                & 0.801                & 0.811                & 0.847                \\
Flan-T5-Enc-Large    & 0.858                & 0.912                & 0.884                & 0.780                & 0.694                & 0.734                & 0.767                & 0.719                & 0.742                & 0.802                & 0.775                & 0.787                & 0.831                \\
Flan-T5-Enc-XL       & 0.851                & 0.927                & 0.889                & 0.798                & 0.694                & 0.742                & 0.840                & 0.656                & 0.737                & 0.830                & 0.759                & 0.789                & 0.838                \\
\rowcolor{lightgray}
\multicolumn{14}{l}{\textsc{Instruction Finetuned encoder-decoder Models}} \\
Flan-T5-Large & 0.907                & 0.883                & 0.901                & 0.766                & 0.837                & 0.800                & 0.800                & 0.750                & 0.774                & 0.824                & 0.823                & 0.825                & 0.861                \\
Flan-T5-XL   & 0.900                & 0.888                & 0.894                & 0.775                & 0.791                & 0.783                & 0.788                & 0.813                & 0.800                & 0.821                & 0.830                & 0.825                & 0.853                \\
\rowcolor{lightgray}
\multicolumn{14}{l}{\textsc{Prompting Methods}} \\
Flan-T5-XXL Zero-shot & 0.792                & 0.726                & 0.758                & 0.435                & 0.378                & 0.404                & 0.244                & 0.688                & 0.361                & 0.491                & 0.597                & 0.508                & 0.612                \\
Flan-UL2 Zeroshot    & 1.000                & 0.003                & 0.005                & 0.133                & 0.133                & 0.133                & 0.075                & 0.969                & 0.139                & 0.403                & 0.368                & 0.092                & 0.095                \\
\midrule
Interpretations & \multicolumn{3}{c}{No Intent} & \multicolumn{3}{c}{Weak Intent} & \multicolumn{3}{c}{Strong Intent} & \multicolumn{3}{c}{\,} \\
\rowcolor{lightgray}
\multicolumn{14}{l}{\textsc{Prior methods}} \\
\citet{sharma-etal-2020-computational}   & 0.840                & 0.883                & 0.861                & 0.000                & 0.000                & 0.000                & 0.850                & 0.834                & 0.842                & 0.563                & 0.572                & 0.567                & 0.834                \\
\citet{welivita2021large} & 0.874                & 0.852                & 0.863                & 0.000                & 0.000                & 0.000                & 0.843                & 0.853                & 0.848                & 0.572                & 0.568                & 0.570                & 0.826                \\
\citet{li-etal-2023-understanding}       & 0.860                & 0.868                & 0.864                & 0.000                & 0.000                & 0.000                & 0.858                & 0.845                & 0.852                & 0.573                & 0.571                & 0.572                & 0.831                \\
\rowcolor{lightgray}
\multicolumn{14}{l}{\textsc{Encoder-only Models}} \\
BERT-Large           & 0.920                & 0.861                & 0.890                & 0.235                & 0.211                & 0.222                & 0.823                & 0.894                & 0.857                & 0.659                & 0.655                & 0.656                & 0.856                \\
RoBERTa-Large        & 0.952                & 0.774                & 0.854                & 0.154                & 0.316                & 0.207                & 0.805                & 0.932                & 0.864                & 0.637                & 0.674                & 0.641                & 0.828                \\
Flan-T5-Enc-Large    & 0.902                & 0.883                & 0.892                & 0.000                & 0.000                & 0.000                & 0.839                & 0.921                & 0.878                & 0.580                & 0.601                & 0.590                & 0.872                \\
Flan-T5-Enc-XL       & 0.925                & 0.837                & 0.888                & 0.294                & 0.263                & 0.278                & 0.820                & 0.943                & 0.877                & 0.679                & 0.681                & 0.681                & 0.865                \\
\rowcolor{lightgray}
\multicolumn{14}{l}{\textsc{Instruction Finetuned encoder-decoder Models}} \\
Flan-T5-Large & 0.953                & 0.791                & 0.864                & 0.400                & 0.105                & 0.167                & 0.763                & 0.966                & 0.852                & 0.705                & 0.621                & 0.628                & 0.845                \\
Flan-T5-XL    & 0.946                & 0.842                & 0.891                & 0.222                & 0.211                & 0.216                & 0.823                & 0.939                & 0.877                & 0.664                & 0.664                & 0.662                & 0.864                \\
\rowcolor{lightgray}
\multicolumn{14}{l}{\textsc{Prompting Methods}} \\
Flan-T5-XXL Zero-shot & 0.664                & 0.664                & 0.664                & 0.048                & 0.474                & 0.087                & 0.628                & 0.225                & 0.332                & 0.447                & 0.454                & 0.361                & 0.470                \\
Flan-UL2 Zero-shot    & 0.000                & 0.000                & 0.000                & 0.020                & 0.158                & 0.036                & 0.495                & 0.870                & 0.631                & 0.172                & 0.343                & 0.222                & 0.378                \\

\midrule
Explorations & \multicolumn{3}{c}{No Intent} & \multicolumn{3}{c}{Weak Intent} & \multicolumn{3}{c}{Strong Intent} & \multicolumn{3}{c}{\,} \\
\rowcolor{lightgray}
\multicolumn{14}{l}{\textsc{Prior methods}} \\
\citet{sharma-etal-2020-computational}   & 0.979                & 0.974                & 0.976                & 0.783                & 0.643                & 0.706                & 0.723                & 0.825                & 0.771                & 0.828                & 0.814                & 0.818                & 0.945                \\
\citet{welivita2021large} & 0.979                & 0.970                & 0.975                & 0.824                & 0.500                & 0.622                & 0.671                & 0.860                & 0.754                & 0.825                & 0.777                & 0.784                & 0.938                \\
\citet{li-etal-2023-understanding}       & 0.976                & 0.983                & 0.979                & 0.850                & 0.607                & 0.708                & 0.754                & 0.807                & 0.780                & 0.860                & 0.799                & 0.822                & 0.950                \\
\rowcolor{lightgray}
\multicolumn{14}{l}{\textsc{Encoder Models}} \\
BERT-Large           & 0.974                & 0.985                & 0.979                & 0.727                & 0.571                & 0.640                & 0.754                & 0.754                & 0.754                & 0.819                & 0.770                & 0.791                & 0.945                \\
RoBERTa-Large        & 0.972                & 0.985                & 0.974                & 0.857                & 0.643                & 0.735                & 0.772                & 0.772                & 0.772                & 0.867                & 0.800                & 0.828                & 0.950                \\
Flan-T5-Enc-Large    & 0.987                & 0.976                & 0.981                & 0.719                & 0.821                & 0.767                & 0.780                & 0.807                & 0.793                & 0.828                & 0.868                & 0.847                & 0.953                \\
Flan-T5-Enc-XL       & 0.985                & 0.972                & 0.978                & 0.690                & 0.714                & 0.702                & 0.762                & 0.842                & 0.800                & 0.812                & 0.843                & 0.827                & 0.948                \\
\rowcolor{lightgray}
\multicolumn{14}{l}{\textsc{Instruction Finetuned encoder-decoder Models}} \\
Flan-T5-Large & 0.977                & 0.973                & 0.975                & 0.833                & 0.536                & 0.652                & 0.681                & 0.825                & 0.746                & 0.831                & 0.778                & 0.791                & 0.940                \\
Flan-T5-XL    & 0.977                & 0.983                & 0.980                & 0.818                & 0.643                & 0.720                & 0.750                & 0.790                & 0.769                & 0.849                & 0.805                & 0.823                & 0.949                \\
\rowcolor{lightgray}
\multicolumn{14}{l}{\textsc{Prompting Methods}} \\
Flan-T5-XXL Zero-shot & 0.908                & 0.894                & 0.901                & 0.107                & 0.214                & 0.143                & 0.270                & 0.175                & 0.213                & 0.428                & 0.428                & 0.419                & 0.796                \\
Flan-UL2 Zero-shot    & 1.000                & 0.010                & 0.019                & 0.021                & 0.214                & 0.039                & 0.133                & 0.754                & 0.226                & 0.385                & 0.326                & 0.095                & 0.088                \\

\bottomrule
\end{tabular}
}
\label{tab:emh_full_results}
\vspace{-5pt}
\end{table*}

\begin{table*}[t]
\centering
\caption{Full results on Esconv dataset, part 1.}
\vspace{0pt}
\resizebox{\textwidth}{!}{
\begin{tabular}{lrrrrrrrrrr}
\toprule
\, & Precision & Recall & F1 & Precision & Recall & F1 & Macro-Precision & Macro-Recall & Macro-F1 & Accuracy \\

\midrule
Question & \multicolumn{3}{c}{No Intent} & \multicolumn{3}{c}{Have Intent} & \multicolumn{3}{c}{\,} \\
\rowcolor{lightgray}
\multicolumn{11}{l}{\textsc{Prior methods}} \\
\citet{sharma-etal-2020-computational}   & 0.949 & 0.931 & 0.940 & 0.761 & 0.814 & 0.787 & 0.855 & 0.873 & 0.863 & 0.906 \\
\citet{welivita2021large} & 0.956 & 0.917 & 0.936 & 0.732 & 0.842 & 0.783 & 0.844 & 0.880 & 0.860 & 0.901 \\
\citet{li-etal-2023-understanding}       & 0.949 & 0.933 & 0.941 & 0.767 & 0.813 & 0.789 & 0.858 & 0.873 & 0.865 & 0.908 \\

\rowcolor{lightgray}
\multicolumn{11}{l}{\textsc{Encoder Models}} \\
BERT-Large           & 0.949 & 0.941 & 0.945 & 0.788 & 0.815 & 0.801 & 0.869 & 0.878 & 0.873 & 0.914 \\
RoBERTa-Large        & 0.943 & 0.938 & 0.941 & 0.776 & 0.789 & 0.783 & 0.860 & 0.864 & 0.862 & 0.907 \\
Flan-T5-Enc-Large    & 0.952 & 0.928 & 0.940 & 0.758 & 0.828 & 0.791 & 0.855 & 0.878 & 0.866 & 0.907 \\
Flan-T5-Enc-XL       & 0.950 & 0.938 & 0.944 & 0.781 & 0.819 & 0.799 & 0.866 & 0.878 & 0.872 & 0.912 \\

\rowcolor{lightgray}
\multicolumn{11}{l}{\textsc{Instruction Finetuned encoder-decoder Models}} \\
Flan-T5-Large & 0.957 & 0.930 & 0.943 & 0.766 & 0.846 & 0.804 & 0.861 & 0.888 & 0.874 & 0.912 \\
Flan-T5-XL   & 0.956 & 0.929 & 0.942 & 0.763 & 0.844 & 0.802 & 0.860 & 0.887 & 0.872 & 0.911 \\

\rowcolor{lightgray}
\multicolumn{11}{l}{\textsc{Prompting Methods}} \\
Flan-T5-XXL Zero-shot & 0.990 & 0.755 & 0.857 & 0.520 & 0.973 & 0.678 & 0.755 & 0.864 & 0.767 & 0.802 \\
Flan-UL2 Zero-shot    & 0.965 & 0.847 & 0.902 & 0.612 & 0.887 & 0.725 & 0.789 & 0.867 & 0.813 & 0.856 \\

\midrule
Affirmation \& Reassurance & \multicolumn{3}{c}{No Intent} & \multicolumn{3}{c}{Have Intent} & \multicolumn{3}{c}{\,} \\
\rowcolor{lightgray}
\multicolumn{11}{l}{\textsc{Prior methods}} \\
\citet{sharma-etal-2020-computational}  & 0.873 & 0.980 & 0.923 & 0.698 & 0.244 & 0.362 & 0.786 & 0.612 & 0.643 & 0.863 \\
\citet{welivita2021large} & 0.885 & 0.962 & 0.922 & 0.626 & 0.339 & 0.440 & 0.756 & 0.651 & 0.681 & 0.863 \\
\citet{li-etal-2023-understanding}       & 0.886 & 0.959 & 0.921 & 0.614 & 0.344 & 0.441 & 0.750 & 0.652 & 0.681 & 0.862 \\

\rowcolor{lightgray}
\multicolumn{11}{l}{\textsc{Encoder Models}} \\
BERT-Large           & 0.904 & 0.927 & 0.915 & 0.509 & 0.435 & 0.469 & 0.706 & 0.681 & 0.692 & 0.854 \\
RoBERTa-Large        & 0.907 & 0.920 & 0.914 & 0.502 & 0.459 & 0.479 & 0.704 & 0.690 & 0.696 & 0.852 \\
Flan-T5-Enc-Large    & 0.874 & 0.982 & 0.925 & 0.642 & 0.187 & 0.290 & 0.758 & 0.585 & 0.607 & 0.864 \\
Flan-T5-Enc-XL      & 0.874 & 0.982 & 0.925 & 0.642 & 0.187 & 0.290 & 0.758 & 0.585 & 0.607 & 0.864 \\

\rowcolor{lightgray}
\multicolumn{11}{l}{\textsc{Instruction Finetuned encoder-decoder Models}} \\
Flan-T5-Large & 0.914 & 0.907 & 0.910 & 0.529 & 0.550 & 0.538 & 0.721 & 0.728 & 0.724 & 0.850 \\
Flan-T5-XL    & 0.911 & 0.920 & 0.916 & 0.555 & 0.525 & 0.540 & 0.733 & 0.723 & 0.728 & 0.858 \\

\rowcolor{lightgray}
\multicolumn{11}{l}{\textsc{Prompting Methods}} \\
Flan-T5-XXL Zero-shot & 0.866 & 0.945 & 0.904 & 0.442 & 0.230 & 0.303 & 0.654 & 0.588 & 0.603 & 0.831 \\
Flan-UL2 Zero-shot    & 0.924 & 0.463 & 0.617 & 0.220 & 0.799 & 0.345 & 0.572 & 0.631 & 0.481 & 0.516 \\

\midrule
Providing Suggestions & \multicolumn{3}{c}{No Intent} & \multicolumn{3}{c}{Have Intent} & \multicolumn{3}{c}{\,} \\
\rowcolor{lightgray}
\multicolumn{11}{l}{\textsc{Prior methods}} \\
\citet{sharma-etal-2020-computational}   & 0.897 & 0.949 & 0.922 & 0.619 & 0.433 & 0.510 & 0.758 & 0.691 & 0.716 & 0.865 \\
\citet{welivita2021large} & 0.920 & 0.912 & 0.916 & 0.563 & 0.587 & 0.575 & 0.741 & 0.750 & 0.745 & 0.859 \\
\citet{li-etal-2023-understanding}       & 0.912 & 0.927 & 0.919 & 0.585 & 0.537 & 0.560 & 0.749 & 0.732 & 0.740 & 0.864 \\

\rowcolor{lightgray}
\multicolumn{11}{l}{\textsc{Encoder Models}} \\
BERT-Large           & 0.911 & 0.923 & 0.917 & 0.584 & 0.545 & 0.564 & 0.748 & 0.734 & 0.740 & 0.861 \\
RoBERTa-Large        & 0.937 & 0.884 & 0.910 & 0.543 & 0.698 & 0.611 & 0.740 & 0.791 & 0.760 & 0.853 \\
Flan-T5-Enc-Large    & 0.909 & 0.941 & 0.925 & 0.634 & 0.521 & 0.572 & 0.772 & 0.731 & 0.748 & 0.872 \\
Flan-T5-Enc-XL       & 0.932 & 0.911 & 0.921 & 0.593 & 0.660 & 0.625 & 0.762 & 0.786 & 0.773 & 0.870 \\

\rowcolor{lightgray}
\multicolumn{11}{l}{\textsc{Instruction Finetuned encoder-decoder Models}} \\
Flan-T5-Large & 0.934 & 0.908 & 0.921 & 0.580 & 0.663 & 0.619 & 0.757 & 0.786 & 0.770 & 0.869 \\
Flan-T5-XL    & 0.924 & 0.927 & 0.925 & 0.608 & 0.597 & 0.603 & 0.766 & 0.762 & 0.764 & 0.874 \\

\rowcolor{lightgray}
\multicolumn{11}{l}{\textsc{Prompting Methods}} \\
Flan-T5-XXL Zero-shot & 0.963 & 0.718 & 0.823 & 0.366 & 0.856 & 0.513 & 0.665 & 0.787 & 0.668 & 0.740 \\
Flan-UL2 Zero-shot    & 0.975 & 0.474 & 0.638 & 0.253 & 0.936 & 0.399 & 0.614 & 0.705 & 0.518 & 0.548 \\

\midrule
Information & \multicolumn{3}{c}{No Intent} & \multicolumn{3}{c}{Have Intent} & \multicolumn{3}{c}{\,} \\
\rowcolor{lightgray}
\multicolumn{11}{l}{\textsc{Prior methods}} \\
\citet{sharma-etal-2020-computational}   & 0.951 & 0.990 & 0.970 & 0.492 & 0.155 & 0.235 & 0.722 & 0.573 & 0.603 & 0.943 \\
\citet{welivita2021large} & 0.953 & 0.979 & 0.966 & 0.365 & 0.203 & 0.261 & 0.659 & 0.591 & 0.613 & 0.935 \\
\citet{li-etal-2023-understanding}       & 0.954 & 0.964 & 0.959 & 0.282 & 0.237 & 0.257 & 0.618 & 0.600 & 0.608 & 0.922 \\

\rowcolor{lightgray}
\multicolumn{11}{l}{\textsc{Encoder-only Models}} \\
BERT-Large           & 0.954 & 0.974 & 0.964 & 0.385 & 0.259 & 0.310 & 0.669 & 0.616 & 0.637 & 0.931 \\
RoBERTa-Large        & 0.958 & 0.967 & 0.962 & 0.385 & 0.327 & 0.354 & 0.671 & 0.647 & 0.658 & 0.928 \\
Flan-T5-Enc-Large    & 0.961 & 0.946 & 0.953 & 0.314 & 0.391 & 0.348 & 0.637 & 0.668 & 0.651 & 0.912 \\
Flan-T5-Enc-XL       & 0.961 & 0.946 & 0.953 & 0.314 & 0.391 & 0.348 & 0.637 & 0.668 & 0.651 & 0.912 \\

\rowcolor{lightgray}
\multicolumn{11}{l}{\textsc{Instruction Finetuned encoder-decoder Models}} \\
Flan-T5-Large & 0.956 & 0.977 & 0.967 & 0.398 & 0.251 & 0.308 & 0.677 & 0.614 & 0.637 & 0.936 \\
Flan-T5-XL    & 0.956 & 0.985 & 0.970 & 0.490 & 0.246 & 0.328 & 0.723 & 0.616 & 0.649 & 0.943 \\

\rowcolor{lightgray}
\multicolumn{11}{l}{\textsc{Prompting Methods}} \\
Flan-T5-XXL Zero-shot & 0.963 & 0.877 & 0.918 & 0.175 & 0.438 & 0.250 & 0.569 & 0.658 & 0.584 & 0.852 \\
Flan-UL2 Zero-shot    & 0.962 & 0.834 & 0.893 & 0.139 & 0.448 & 0.212 & 0.550 & 0.641 & 0.553 & 0.812 \\

\bottomrule
\end{tabular}
}
\label{tab:esconv_full_results_p1}
\vspace{-5pt}
\end{table*}

\begin{table*}[t]
\centering
\caption{Full results on Esconv dataset, part 2.}
\vspace{0pt}
\resizebox{\textwidth}{!}{
\begin{tabular}{lrrrrrrrrrr}
\toprule
\, & Precision & Recall & F1 & Precision & Recall & F1 & Macro-Precision & Macro-Recall & Macro-F1 & Accuracy \\

\midrule
Self-disclosure & \multicolumn{3}{c}{No Intent} & \multicolumn{3}{c}{Have Intent} & \multicolumn{3}{c}{\,} \\
\rowcolor{lightgray}
\multicolumn{11}{l}{\textsc{Prior methods}} \\
\citet{sharma-etal-2020-computational}   & 0.952 & 0.981 & 0.966 & 0.741 & 0.527 & 0.616 & 0.846 & 0.754 & 0.791 & 0.937 \\
\citet{welivita2021large} & 0.965 & 0.968 & 0.966 & 0.685 & 0.663 & 0.674 & 0.825 & 0.815 & 0.820 & 0.939 \\
\citet{li-etal-2023-understanding}       & 0.957 & 0.978 & 0.967 & 0.734 & 0.579 & 0.647 & 0.845 & 0.779 & 0.807 & 0.940 \\

\rowcolor{lightgray}
\multicolumn{11}{l}{\textsc{Encoder-only Models}} \\

BERT-Large           & 0.963 & 0.962 & 0.962 & 0.625 & 0.631 & 0.628 & 0.794 & 0.796 & 0.795 & 0.932 \\
RoBERTa-Large        & 0.963 & 0.965 & 0.964 & 0.646 & 0.634 & 0.640 & 0.804 & 0.799 & 0.802 & 0.935 \\
Flan-T5-Enc-Large    & 0.974 & 0.951 & 0.962 & 0.605 & 0.744 & 0.668 & 0.789 & 0.848 & 0.815 & 0.932 \\
Flan-T5-Enc-XL       & 0.970 & 0.957 & 0.964 & 0.626 & 0.708 & 0.665 & 0.798 & 0.833 & 0.814 & 0.935 \\

\rowcolor{lightgray}
\multicolumn{11}{l}{\textsc{Instruction Finetuned encoder-decoder Models}} \\
Flan-T5-Large & 0.975 & 0.962 & 0.968 & 0.673 & 0.758 & 0.713 & 0.824 & 0.860 & 0.840 & 0.943 \\
Flan-T5-XL    & 0.969 & 0.970 & 0.970 & 0.709 & 0.705 & 0.707 & 0.839 & 0.837 & 0.838 & 0.945 \\

\rowcolor{lightgray}
\multicolumn{11}{l}{\textsc{Prompting Methods}} \\
Flan-T5-XXL Zero-shot & 0.959 & 0.842 & 0.897 & 0.300 & 0.652 & 0.411 & 0.630 & 0.747 & 0.654 & 0.824 \\
Flan-UL2 Zero-shot    & 0.962 & 0.774 & 0.858 & 0.246 & 0.708 & 0.365 & 0.604 & 0.741 & 0.612 & 0.768 \\
\midrule
Restatement or Paraphrase & \multicolumn{3}{c}{No Intent} & \multicolumn{3}{c}{Have Intent} & \multicolumn{3}{c}{\,} \\
\rowcolor{lightgray}
\multicolumn{11}{l}{\textsc{Prior methods}} \\
\citet{sharma-etal-2020-computational}   & 0.952 & 0.996 & 0.974 & 0.790 & 0.222 & 0.346 & 0.871 & 0.609 & 0.660 & 0.949 \\
\citet{welivita2021large} & 0.960 & 0.992 & 0.976 & 0.750 & 0.367 & 0.492 & 0.855 & 0.679 & 0.734 & 0.954 \\
\citet{li-etal-2023-understanding}       & 0.960 & 0.988 & 0.974 & 0.661 & 0.362 & 0.468 & 0.811 & 0.675 & 0.721 & 0.950 \\

\rowcolor{lightgray}
\multicolumn{11}{l}{\textsc{Encoder-only Models}} \\
BERT-Large           & 0.961 & 0.983 & 0.972 & 0.613 & 0.411 & 0.492 & 0.787 & 0.697 & 0.732 & 0.947 \\
RoBERTa-Large        & 0.962 & 0.982 & 0.972 & 0.611 & 0.429 & 0.504 & 0.787 & 0.705 & 0.738 & 0.947 \\
Flan-T5-Enc-Large    & 0.961 & 0.991 & 0.976 & 0.742 & 0.398 & 0.518 & 0.851 & 0.695 & 0.747 & 0.953 \\
Flan-T5-Enc-XL       & 0.961 & 0.991 & 0.976 & 0.742 & 0.398 & 0.518 & 0.851 & 0.695 & 0.747 & 0.953 \\

\rowcolor{lightgray}
\multicolumn{11}{l}{\textsc{Instruction Finetuned encoder-decoder Models}} \\
Flan-T5-Large & 0.964 & 0.990 & 0.977 & 0.729 & 0.429 & 0.540 & 0.846 & 0.709 & 0.758 & 0.956 \\
Flan-T5-XL    & 0.962 & 0.992 & 0.977 & 0.763 & 0.397 & 0.523 & 0.863 & 0.695 & 0.750 & 0.956 \\

\rowcolor{lightgray}
\multicolumn{11}{l}{\textsc{Prompting Methods}} \\
Flan-T5-XXL Zero-shot & 0.972 & 0.828 & 0.894 & 0.192 & 0.630 & 0.294 & 0.582 & 0.729 & 0.594 & 0.816 \\
Flan-UL2 Zero-shot    & 0.973 & 0.195 & 0.325 & 0.069 & 0.918 & 0.128 & 0.521 & 0.556 & 0.226 & 0.239 \\

\midrule
Reflection of Feelings & \multicolumn{3}{c}{No Intent} & \multicolumn{3}{c}{Have Intent} & \multicolumn{3}{c}{\,} \\
\rowcolor{lightgray}
\multicolumn{11}{l}{\textsc{Prior methods}} \\
\citet{sharma-etal-2020-computational}   & 0.935 & 0.998 & 0.965 & 0.813 & 0.099 & 0.177 & 0.874 & 0.549 & 0.571 & 0.934 \\
\citet{welivita2021large} & 0.941 & 0.991 & 0.966 & 0.639 & 0.202 & 0.307 & 0.790 & 0.597 & 0.636 & 0.934 \\
\citet{li-etal-2023-understanding}       & 0.944 & 0.979 & 0.961 & 0.485 & 0.252 & 0.332 & 0.715 & 0.616 & 0.649 & 0.927 \\

\rowcolor{lightgray}
\multicolumn{11}{l}{\textsc{Encoder-only Models}} \\
BERT-Large           & 0.940 & 0.978 & 0.958 & 0.487 & 0.254 & 0.333 & 0.713 & 0.616 & 0.646 & 0.922 \\
RoBERTa-Large        & 0.923 & 1.000 & 0.960 & 0.000 & 0.000 & 0.000 & 0.461 & 0.500 & 0.480 & 0.923 \\
Flan-T5-Enc-Large    & 0.942 & 0.976 & 0.959 & 0.506 & 0.289 & 0.368 & 0.724 & 0.633 & 0.663 & 0.923 \\
Flan-T5-Enc-XL       & 0.949 & 0.961 & 0.955 & 0.450 & 0.380 & 0.412 & 0.699 & 0.671 & 0.684 & 0.916 \\

\rowcolor{lightgray}
\multicolumn{11}{l}{\textsc{Instruction Finetuned encoder-decoder Models}} \\
Flan-T5-Large & 0.949 & 0.966 & 0.957 & 0.427 & 0.330 & 0.372 & 0.688 & 0.648 & 0.665 & 0.920 \\
Flan-T5-XL    & 0.946 & 0.978 & 0.962 & 0.493 & 0.279 & 0.356 & 0.720 & 0.628 & 0.659 & 0.928 \\

\rowcolor{lightgray}
\multicolumn{11}{l}{\textsc{Prompting Methods}} \\
Flan-T5-XXL Zero-shot & 0.971 & 0.514 & 0.672 & 0.113 & 0.798 & 0.198 & 0.542 & 0.656 & 0.435 & 0.535 \\
Flan-UL2 Zero-shot    & 0.964 & 0.637 & 0.767 & 0.129 & 0.694 & 0.217 & 0.547 & 0.666 & 0.492 & 0.641 \\
\bottomrule
\end{tabular}
}
\label{tab:esconv_full_results_p2}
\vspace{-5pt}
\end{table*}

\end{document}